\documentclass[dvips, preprint]{imsart}

\usepackage{amsthm,amsmath}

\numberwithin{equation}{section}
\usepackage{graphicx}
\usepackage{dcolumn}
\usepackage{bm}
\usepackage{verbatim}
\usepackage{ulem}
\usepackage{color}
\usepackage[usenames,dvipsnames]{xcolor}

\usepackage{hyperref}
\hypersetup{breaklinks=true,
pagecolor=white,
colorlinks=true}
\makeatletter
\def\url@leostyle{%
  \@ifundefined{selectfont}{\def\UrlFont{\sf}}{\def\UrlFont{\scriptsize\ttfamily}}}
\makeatother
\renewcommand{\a}{\displaystyle}

\makeatletter
\newcommand{\thickline}{%
    \noalign {\ifnum 0=`}\fi \hrule height 1pt
    \futurelet \reserved@a \@xhline
}
\newcolumntype{"}{@{\hskip\tabcolsep\vrule width 1pt\hskip\tabcolsep}}
\makeatother

\begin{document}

\begin{frontmatter}

\title{Finding Important Genes from High-Dimensional Data: An Appraisal of Statistical
Tests and Machine-Learning Approaches}
\runtitle{Casual Inference from High-Dimensional Data}

\begin{aug}

\author{\fnms{Chamont} \snm{Wang}\ead[label=e1]{wang@tcnj.edu}} 
\hspace{-0.07in} \and
\author{\fnms{Jana} \snm{Gevertz}\ead[label=e2]{gevertz@tcnj.edu}}
\affiliation{The College of New Jersey}

 \author{\fnms{Chaur-Chin} \snm{Chen}\ead[label=e3]{daniel.ccchen@gmail.com}}
 \affiliation{National Tsing-Hua University}

 \author{\fnms{Leonardo} \snm{Auslender}\ead[label=e4]{leoldv12@gmail.com}}
 \affiliation{Cisco Systems, Inc.}

 \address{Chamont Wang is Professor of Mathematics and Statistics at The College of New
 Jersey, and Jana Gevertz is Assistant Professor of Mathematics and Statistics at The
 College of New Jersey, P. O. Box 7718 Ewing NJ 08628 USA \printead{e1,e2}.}

\address{Chaur-Chin Chen is Professor of Computer Science at National
 Tsing-Hua University, Kuang-Fu Road, Hsinchu, Taiwan 30013 \printead{e3}.}

\address{Leonardo Auslender is Senior Consultant at Cisco Systems, Inc.,
USA \printead{e4}.}
\end{aug}

\runauthor{C. Wang et al.}

\begin{abstract} 

Over the past decades, statisticians and machine-learning researchers have developed literally thousands of new tools for the reduction of high-dimensional data in order to identify the variables most responsible for a particular trait.  These tools have applications in a plethora of settings, including data analysis in the fields of business, education, forensics, and biology (such as microarray, proteomics, brain imaging), to name a few.

In the present work, we focus our investigation on the limitations and potential misuses of certain tools in the analysis of the benchmark colon cancer data (2,000 variables; Alon et al., 1999) and the prostate cancer data (6,033 variables; Efron, 2010, 2008).  Our analysis demonstrates that models that produce 100\% accuracy measures often select different sets of genes and cannot stand the scrutiny of parameter estimates and model stability.

Furthermore, we created a host of simulation datasets and ``artificial diseases" to evaluate the reliability of commonly used statistical and data mining tools.  We found that certain widely used models can classify the data with 100\% accuracy without using any of the variables responsible for the disease.  With moderate sample size and suitable pre-screening, stochastic gradient boosting will be shown to be a superior model for gene selection and variable screening from high-dimensional datasets.

\end{abstract}



\end{frontmatter}

\section{Introduction}

High-dimensional data is increasingly common in modern statistical analysis, where the
number of variables is on the order of thousands and beyond.  In an international competition on the analysis of breast cancer, the raw data has $p = 32,670$ bins for predictors (Hand, 2008).  At the Center for Cancer Research, the proteomic data for ovarian cancer has $p = 360,000$ predictors and it is free for anyone to download (\url{http://home.ccr.cancer.gov/ncifdaproteomics/OvarianCD_PostQAQC.zip}).  In a bankruptcy application, Foster and Stine (2004, {\it JASA}) used $p = 67,000$ predictors in their model.  Efron (2008, 2010) mentioned
$p = 6,000$ for microarray gene expression data, $p = 15,445$ for imaging processing, and
$p > 500,000$ for SNP analysis.  In a different area of application, a \textit{New York Times}
article (October 30, 2005) reported that Google utilizes millions of variables about its
users and advertisers in its predictive modeling to deliver the message to which each user
is most likely to respond.  Furthermore, in the field of astrostatistical applications, Efron (2008)
mentioned $p = 10^{10}$, which would dwarf other data and is probably qualified to be
called Mother of High-Dimensional Data.

In certain areas of \textit{predictive modeling} with high-dimensional data, statistical methods and
machine-learning tools appear to be doing very well, but in other applications where \textit{causal
inferences} are involved, the results are often less than satisfactory.  Shmueli (2010,
\textit{Statistical Science}) asked the question, ``To Explain or to Predict?", and emphasized that
``the type of uncertainty associated with explanation is of a different nature than that
associated with prediction" (a la Helmer and Rescher, 1959).  Here explanatory modeling
refers to the statistical analysis of cause-and-effect.  The focus of our study is an
extension of this effort.

For this purpose, we found that the field of gene identification provides an excellent
framework to discuss a number of issues related to the limitations and potential misuses
of causal inference from high-dimensional data.

The datasets in our study are taken from the field of microarray gene identification. This
technology is a powerful tool for measuring the relative expression level of thousands of
genes in a single experiment.  In particular, every cell in an organism expresses its own
set of genes.  Skin cells express different genes than bone cells, and colon cancer cells
express different genes than normal colon cells.  Therefore, one way to determine what genes
cause a particular trait or disease is to compare the genes expressed in one cell type to
those expressed in the other cell type.

Microarrays allow this kind of comparative study to take place on a very large scale: the
expression level of thousands of genes can be compared across cell types.  In this study,
our goal is to identify those genes that are differentially expressed in cancer, as these
differentially expressed genes may actually be the cause of cancer formation and
progression.  In order to do this, the hundreds to hundreds of thousands of data points
collected in microarray experiments need to be analyzed using sound and robust statistical
methods.

Given the massive size of the datasets, and the number of statistical techniques available
for analysis, the field has attracted an enormous amount of attention from researchers
around the globe.  A survey of the literature reveals that there is a huge variety of
techniques for selecting genes whose aberrant expression correlates with a particular
tissue type or disease state.  These techniques can be roughly grouped into the following
two categories:

\begin{itemize}
 \item \textit{Multiple hypothesis testing} that includes $t$-tests, the Bonferroni
 correction, false discovery rates, empirical Bayes, Sidak method, $Q$-values, mid
 $p$-values, platform  $p$-value, $F$-test, two-step non-parametric statistical analysis,
 regularized $t$-test, hierarchical lognormal-normal model, etc.  For references, see for
 instance Leek and Storey (2011), Efron (2011, 2010, 2008), Bar et al. (2010), Storey
 (2010), Ferreira1 and Zwinderman (2006), Dudoit, Shaffer and Boldrick (2003), Sierra
 and Echeverria (2003), Guyon and Elisseeff (2003), Benjamini and Hochberg (1995), etc.

 \item \textit{Statistical models and machine-learning methods} that includes logistic
 regression, ANOVA, support vector machines, neural networks, random forests, $k$-nearest
 neighbors, diagonal linear discriminant analysis, na\"{i}ve Bayes, nearest centroid, rough
 set, emerging pattern, a genetic-algorithm-based Fisher's discriminant analysis,
 Mahalanobis decorrelation, latent class analysis, Laplace approximated EM microarray
 analysis, pathway analysis, neighborhood mutual information, fuzzy mutual information,
 and numerous other variations.  For references, see for instance Huang et al. (2011),
 Zuber and Strimmer (2011), Wang and Simon (2011), Bar et al. (2010), Hu et al. (2010),
 Mongan et al. (2010), Cordell (2009), Lee et al. (2008), Dean and Raftery (2008), Ma
 and Huang (2007), Guyon and Elisseeff (2003), etc.

\end{itemize}

In addition, there are countless references within each of the above categories.
Furthermore, each technique in the above lists can have endless variations.  For
instance,

\begin{itemize}
 \item In their paper, ``Should We Abandon the $t$-test in the Analysis of Gene
 Expression Microarray Data," Jeanmougin et al. (2010) considered eight different tests
 representative of various modeling strategies in gene expression data: ANOVA
 (homoscedastic), Welch's $t$-test (heteroscedastic), RVM (homoscedastic), limma
 (homoscedastic and based on a Bayesian framework) and SMVar (heteroscedastic and based
 on structural model), plus two non-parametric approaches with the Wilcoxon's test and
 the SAM test.

 \item Regression-based methods would include sliced inverse regression, correlated
 component regression, lasso regression, the elastic net, non-negative garrote method,
 etc.   Among these methods, in the area of lasso regression, there are nine different
 methods in a MATLAB toolbox by Liu et al. (2009), and for each method one can define
 the penalty functions in multiple ways to generate new models
 (\url{http://www.public.asu.edu/~jye02/Software/SLEP/}).  It is conceivable that one
 may find countless other variations in the literature.

 \item In the area of support vector machines (SVM), there are at least 25 different
 kernels

 (\url{http://crsouza.blogspot.com/2010/03/kernel-functions-for-machine-learning.html}).

 With some modifications and hybridizations, it would be straightforward to generate
 hundreds of additional kernels without knowing which one is best suited to analyze
 the data at hand.

 \item The well-known CART method, classification and regression tree, as laid out in
 Breiman, Friedman, Olshen, and Stone (1983) has endless variations in the
 machine-leaning literature including Gini index, chi-square criterion, entropy,
 genetic-algorithm tree (Cha and Tappert, 2009), and neural-network tree (SAS, 2003).
 The techniques can easily fill up a huge volume to greet a biologist or to send
 him/her down the wrong path when it comes to analyzing a high-dimensional dataset.

 \item Neural Network models have even more variations than CART: multilayer
 perceptrons (MLPs), radial basis function (RBF) networks, and many other forms of
 network architectures.  SAS, a software package, provides nine different kinds of
 architectures, fourteen different kinds of error functions, eight different kinds
 of combination functions, and twelve different kinds of activation functions. There
 are a lot more in scholarly publications (see e.g., a 2009 book by Pereira and Rao
 titled, \textit{Data Mining using Neural Networks: A Guide for Statisticians}). A Google
 search of ``neural network architecture" (with quotations marks) rendered 3,130,000
 results, with more coming every day.

\end{itemize}

The situation reminds us the famous example from 1972 where 10,465 techniques were
constructed in the estimation of a statistical quantity called the location parameter
(Stigler, 2010, a la Andrews et al., 1972). For the modern-day gene hunt, the number
of techniques available is equally endless.

A natural question is: how reliable are these statistical tests and modeling
techniques?  Specifically, one may ask whether the models are stable, whether they
are consistent, and whether it is true that ``the increased level of algorithmic
complexity does not always translate to improved biological understanding" (Mongan
et al., 2010).  Along the same line of inquiry, Wang and Simon (2011, p. 22, Table
5) found that many tools achieved high prediction accuracies, yet did so using
different important genes for the same disease.   In an earlier study, Efron (2008,
p. 7) pointed out that

\begin{quote}
  The prostate data has E(Fdr) = 0.68, indicating low power [here E(Fdr) = the
  expected value of false discovery rate]. If the whole study were rerun, we
  could expect a different list of 50 likely nonnull genes, barely overlapping
  with the first list.
\end{quote}

In short, the scientific literature focused on the identification of relevant
genes from microarray data is vast and not necessarily reliable. Consequently, the
main objective of this article is to evaluate some widely-used statistical
tests and machine-learning approaches in the analysis of microarray data.
Specifically, we look at two microarray datasets (detailed below) and we generate
sets of simulated microarray data for which the genes that contribute to the
diseased state are known. We employed several well-known statistical methods to
identify the differentially expressed (important) genes and classify the datasets.
Based on the results for the experimental and simulated datasets, we make
recommendations on the statistical methods we find to be the most reliable.

Our study was motivated by two datasets that are widely known in the field
of cancer research:

\begin{itemize}

 \item Colon cancer from Alon et al., (1999).  The data consists of 2,000 genes
 and 62 patients, 40 who have colon cancer, and 22 who do not.

 \item Prostate cancer from Singh et al. (2002).  The data consists of 6,033
 genes and 102 patients,  52 who have cancer and 50 who are healthy.

\end{itemize}

In Section 2, we review previous results from the colon cancer data and then
present new results on other models that outperform the original results in terms
of accuracy and the number of genes selected.  Furthermore, we discuss the merits
and potential pitfalls of the top models we tested on both the colon and prostate
cancer data, cautioning that even the results from these top models may be
deceiving under certain conditions.  In Section 3, the top models undergo further
scrutiny when we test their performance in a variety of scenarios using simulated
data.  The results of analyzing the simulated data further strengthen our belief
that several popular models may mislead investigators analyzing microarray data.
In Section 4, the relationship between sample size, number of genes and
statistical reliability is explored in depth. Our analysis suggests that gradient
boosting is a significantly better tool than the others explored, and that the
sample size used in some microarray experiments is not sufficiently large for
most statistical methods to accurately and consistently select the most
important genes to classify the data.

\section{The Reliability of Statistical  Methods (I): Results from Real Datasets}

\subsection{Three Statistical Methods for Analyzing Microarray Data}

In this Section, we review previous statistical analyses of the colon cancer
dataset.  We then compare the performances of our models with these
previously-published results.  Much to our delight, some of our models achieved
100\% accuracy in classifying the data in multiple runs
with different random splits of the data into training and validation purposes.
In addition, our models achieved this accuracy using fewer genes than most
previously-published analyses.  However, our further investigations indicate
that the models are not reliable, as will be seen in the subsequent discussions.

Table 1 (see next page) presents a brief summary of the previous analysis of the
colon cancer dataset.  We note that the models attempt to classify a sample as
cancerous or normal using anywhere from 5 to 2,000 genes, and the error rates range
from 11.3\% to 34\%.

\begin{table}[ht!]
\caption{A brief summary of the journal results on the colon cancer data.}
\label{Table 1}
\begin{tabular}{llll}
\thickline
 & \textbf{Variable} & \textbf{\# of Genes} & \textbf{Prediction} \\
 & \textbf{Selection} & \textbf{Selected} & \textbf{Error} \\ \thickline
Blind Bet (No Model) & - - & 2000 & 33\% \\ \hline

Alon et al. (1999) & &  \\
\textit{Proc. Natl. Acad. Sci} & clustering & 500 & n/a \\ \hline

Weston et al. (2001) &  &   \\
\textit{Adv Neural Informat} & SVM & 15 & 11.4\% \\ \hline

Guyon et al. (2002) & & \\
\textit{Machine Learning} & SVM & 8 & 34\% \\ \hline

Weston et al. (2003) & &  \\
\textit{J. Machine Learning} & kernel methods & 20 & 13.7\% \\ \hline

Su et al. (2002) & &  \\
\textit{Bioinformatics} & $t$-tests, SVM & 100 & n/a \\ \hline

Do et al. (2005) & &  \\
\textit{J. Royal Stat Soc.} & Fdr & 1938 & n/a \\ \hline

Ma et al. (2007) & &  \\
\textit{BMC Bioinformatics} & Lasso & 19 & 12.9\% \\ \hline

Lee et al. (2008) & &  \\
\textit{J. Biopharmaceut. Stat} & SVM (1-norm) & 8 & 11.3\%$^*$ \\ \hline

Lee et al. (2008) & &  \\
\textit{J. Biopharmaceut. Stat} & SVM (IFFS) & 5 & 11.3\%$^*$ \\ \hline

Bar et al. (2010) & &  \\
\textit{Statistical Science} & Laplace EM & 61 & n/a \\ \thickline


\end{tabular}
\end{table}

The first statistical method we used to analyze the colon cancer microarray data
was that of partial least squares (PLS) with leave-one-out cross-validation.
Here leave-one-out means that, given $n$ observations, the model was trained
using $n-1$ data points, and the model was used to predict whether the remaining
data is representative of cancer or no cancer.  The model was run $n$ times by
altering which $n-1$ of the $n$ data points were used for training, and which
dataset was being classified as cancerous or normal based on the trained model.
Quite impressively, when PLS selected the nine most important genes, the
leave-one-out prediction error was 0\% (Table 2, see next page).  This indicates
that the model could always correctly classify the one microarray dataset that was
not used for training purposes.

Before running the PLS models to get the data shown in Table 2, the original 2,000
genes in the dataset were prescreened by an $R$-square variable selection procedure
which selected only 25 genes from the larger pool.  The $R$-square procedure is one
of many prescreening techniques available for reducing the dimensionality of a
dataset before searching for the most important genes (Guyon and Elisseeff, 2003).
This prescreening procedure is essential, as PLS does not perform reliably using
thousands of predictors at a time.  The genes that were selected from the $R$-square
variable selection procedure are then used in the subsequent runs of the partial
least squares model.  The default PLS with 25 genes achieved 0\% error rate, meaning
it always correctly classified whether the remaining data set represented cancer or
no cancer.  We then used a stepwise elimination process to cut the low-ranking genes
from the model.  We found that the 16 lowest-ranking genes of the 25 prescreened
genes could be eliminated from the model without impacting predictions: that is,
using only the 9 genes whose expression varies the most between cancer and no cancer,
the error rate of PLS is 0\%.  However, if we cut down to the top 8 genes, the error
rate went up slightly to 3.2\%. \newpage

\begin{table}[ht!]
\caption{PLS achieved 0\% error rate on the colon cancer
data with 9 genes (PLS-1).  If we cut the least important gene from the list of 9
to get 8 genes (PLS-2), the error rate goes up to 3.2\%, still significantly better
than those in Table 1.}
\label{Table 2}
\begin{tabular}{lcc}
\thickline
 & \textbf{\# of Genes} & \textbf{Leave-One-Out} \\
 & \textbf{Selected} & \textbf{Prediction Error} \\
\thickline
 PLS-1 & 9 & 0\% \\
 PLS-2 & 8 & 3.2\% \\ \thickline
\end{tabular}
\end{table}

The next statistical methods we utilized to analyze the colon cancer data are that
of logistic regression and neutral networks (NN), both of which require variable
prescreening similar to PLS.  Our regression and neural network analyses are based
on random splits of the data into a training set (75\% of the data) and validation
set (25\% of the data).  The results of analyzing the colon cancer data using these
statistical methods are displayed in Table 3.

\begin{table}[ht!]
\caption{A comparison of model performances with backward elimination (training data:
75\%, validation data: 25\%).}
\label{Table 3: colon_model_comparison}
\begin{tabular}{lcrr}
\thickline
 & \textbf{\# of} & \textbf{Training} & \textbf{Validation} \\
 & \textbf{Genes} & \textbf{ Error Rate} & \textbf{Error Rate} \\ \thickline
 Regression-1 & 13 & 0\% & 0\% \\
 Regression-2 & 12 & 0\% & 5.9\% \\
 PLS-3 & 13 & 0\% & 0\% \\
 PLS-4 & 12 & 0\% & 5.9\% \\
 Neural Network-1 & 19 & 0\% & 0\% \\
 Neural Network-2 & 18 & 4.3\% & 13.3\% \\ \thickline
\end{tabular}
\end{table}

From Tables 2 and 3, the best model appears to be PLS-1 with 9 genes and 0\%
leave-one-out error rate.  While the error rate is the same as achieved for other
statistical methods, we say PLS-1 appears to be the best model as it required the
fewest number of genes to classify the data with a 0\% error rate.  In order to
facilitate comparison with regression and neural networks, partial least squares was
also conducted by placing 75\% of the data in the training set, and 25\% in the
validation set.  PLS-3 is the analysis when 13 genes were selected via this split of
the data, and PLS-4 is the analysis when 12 genes were selected via this split of
the data.  Notice that when compared to regression and neural networks, PLS with 13
genes does just as well as regression with 13 genes and neural networks with 19 genes.
Notice, however, that when the data is split up in this manner, PLS requires more
than 9 genes to achieve a 0\% error rate.

While the low error rates we have obtained are desirable, this does not demonstrate
that the statistical methods are consistently classifying the data.  For instance,
it is plausible that each method uses a very different set of genes to achieve the
low classification error rates.  In order to explore the between-model consistency,
we looked at the top genes selected by partial least squares, regression and the
neural network (Table 4).  In each case, an $R$-square prescreening method was
applied to reduce the number of genes input into the statistical methods from
2,000 to 38.  Fortunately, the three models appear to be very consistent: seven
genes are selected by all three statistical methods, and three genes are selected
by at least two statistical methods.  Every gene selected by PLS-1 was also selected
by one or both of the other statistical methods.

\begin{table}[ht!]
\caption{Top genes selected by the three different models.  The seven common genes
in the three models are indicated with a *.  The two genes common to regression and
PLS are indicated with a $\dagger$, and the one gene common to regression and neural
network is indicated with a \#.}
\label{Table 4: top_genes_PLS}
\begin{tabular}{llll}
\thickline
 & \textbf{Regression}  & \textbf{PLS-1} & \textbf{Neural Network} \\
 & Accuracy = 100\% & Accuracy = 100\% & Accuracy = 100\% \\
 \thickline
 1 & Gene-1025 & \textcolor{red}{Gene-1769$^*$} & \textcolor{red}{Gene-1769$^*$} \\
 2 & \textcolor{red}{Gene-1231$^*$} & Gene-1466$^\dagger$ & \textcolor{red}{Gene-1231$^*$} \\
 3 & \textcolor{red}{Gene-1351$^*$} & \textcolor{red}{Gene-1367$^*$} & Gene-1421 \\
 4 & \textcolor{red}{Gene-1367$^*$} & Gene-1482$^\dagger$ & Gene-1702 \\
 5 & Gene-1466$^\dagger$ & \textcolor{red}{Gene-419$^*$} & \textcolor{red}{Gene-1351$^*$} \\
 6 & Gene-1482$^\dagger$ & \textcolor{red}{Gene-1351$^*$} & Gene-258 \\
 7 & \textcolor{red}{Gene-1644$^*$} & \textcolor{red}{Gene-1644$^*$} & \textcolor{red}{Gene-1644$^*$} \\
 8 & \textcolor{red}{Gene-1769$^*$} & \textcolor{red}{Gene-249$^*$} & Gene-1475 \\
 9 & Gene-1895$^{\#}$ & \textcolor{red}{Gene-1231$^*$} & Gene-1895$^{\#}$ \\
 10 & \textcolor{red}{Gene-249$^*$} & & \textcolor{red}{Gene-419$^*$} \\
 11 & \textcolor{red}{Gene-419$^*$} & & Gene-1914 \\
 12 & Gene-580 & & \textcolor{red}{Gene-1367$^*$} \\
 13 & Gene-662 & & Gene-1889 \\
 14 & & & Gene-945 \\
 15 & & & \textcolor{red}{Gene-249$^*$} \\ \thickline
\end{tabular}
\end{table}

Having checked for consistency among the statistical methods, we next sought to
ensure that the way the data was being partitioned does not bias the results.  To
undertake this analysis, we used five different random seeds to split the data into
the training set and the validation set.  We found that independent of which datasets
were placed in the training set and which were placed in the validation set, the number
of genes required to achieve a 0\% validation error rate were rather consistent for
each statistical method, as shown in Table 5.

\begin{table}[ht!]
\caption{Five runs of neural network on the colon cancer data with different seeds,
using the same 19 prescreened genes.  In five runs, all neural network trials have
0\% error rate.}
\label{Table 5: colon_NN_seeds}
\begin{tabular}{lccc}
\thickline
 \textbf{Seed} & \textbf{Schwarz}  & \textbf{Training} & \textbf{Validation} \\
 & \textbf{Bayesian Criterion} & \textbf{Error Rate} & \textbf{Error Rate} \\ \thickline
 1 & 178.146 & 0\% & 0\% \\
 2 & 178.159 & 0\% & 0\% \\
 3 & 178.156 & 0\% & 0\% \\
 4 & 178.178 & 0\% & 0\% \\
 5 & 178.148 & 0\% & 0\% \\ \thickline
\end{tabular}
\end{table}

\subsection{Vetting of the Statistical Methods: Should we Believe What we See?}
Taken at face value, our results appear very encouraging in the sense that three
different models, PLS, regression and neural network, achieved 0\% prediction error.
Table 4 also showed that the three models select seven common genes and are very
consistent.  But subsequent analysis in Sections 2.2, 2.3, 3.2, 3.3 will cast doubt
on conclusions drawn from these models.

First of all, from Table 6, we observe that the estimates of the logistic regression
model are relatively small while the corresponding standard errors of the estimates
are quite large.  As a result, the corresponding Wald chi-squares are also very small
and the $p$-values are not significant. Thus, any conclusions drawn from the model are
dubious and may not generalize well for new data.

\begin{table}[ht!]
\caption{Parameter estimates of the regression model.  The estimates are very
small while the corresponding standard errors of the estimates are very large.}
\label{Table 6}
\begin{tabular}{lcccc}
\thickline
\textbf{Gene} &	\textbf{Estimate} &	\textbf{Standard Error}	& \textbf{Wald Chi-Square} &
\textbf{Pr}$\boldsymbol{>}$\textbf{ChiSq}	\\ \thickline
1025	&	0.00608	    &	0.0475	        &	0.02	        &	0.898	\\
1231	&	0.1188	    &	0.4193	        &	0.08	        &	0.777	\\
1351	&	0.0203	    &	0.1179	        &	0.03	        &	0.8631	\\
1367	&	-0.06	&	0.2638	&	0.05	&	0.8202	\\
1466	&	-0.0349	&	0.4738	&	0.01	&	0.9413	\\
1482	&	0.0258	&	0.3734	&	0	&	0.9449	\\
1644	&	0.0523	&	0.29	&	0.03	&	0.857	\\
1769	&	-0.1642	&	0.5339	&	0.09	&	0.7584	\\
1895	&	-0.00606	&	0.0785	&	0.01	&	0.9385	\\
249	&	0.00391	&	0.0111	&	0.12	&	0.7249	\\
419	&	0.04	&	0.1262	&	0.1	&	0.751	\\
580	&	-0.0267	&	0.1666	&	0.03	&	0.8726	\\
662	&	-0.0103	&	0.0406	&	0.06	&	0.799	\\ \thickline

\end{tabular}
\end{table}

The problem is related to ``complete separation" in binary regression where the maximum
likelihood function does not exist and the iterations do not converge.  As a result, the
model, albeit with 0\% error rates in multiple runs on different seeds, may not hold up to
future observations.  A discussion of this phenomenon can be found at
\url{http://www.ats.ucla.edu/stat/sas/library/logistic.pdf}.  Another reference on the
convergence of the maximum likelihood estimate is in Stokes (2004).  Potential ways to fix
this problem for logistic regression have been proposed; see, e.g., Firth (1993), Heinze
and Schemper (2002), and Park and Hastie (2008).

Next, we considered the parameter estimates of PLS as shown in Table 7.

\begin{table}[ht!]
\caption{Parameter estimates of the PLS model.  Standardized Parameter Estimates are very
small so this model may not generalize well to new data.}
\label{Table 7}
\begin{tabular}{lcc}
\thickline
\textbf{Gene} &	\textbf{Standardized Parameter} & \textbf{Rejected by Parameter} \\
 & \textbf{Estimate} & \textbf{Estimate?} \\ \thickline
1769	&	-0.69336	&	No	\\
1466	&	-0.31386	&	No	\\
1367	&	-0.29207	&	No	\\
1482	&	0.1661	&	No	\\
419	&	0.30232	&	No	\\
1351	&	0.3035	&	No	\\
1644	&	0.37666	&	No	\\
249	&	0.45373	&	No	\\
1231	&	0.50063	&	No	\\ \thickline
\end{tabular}
\end{table}

\noindent The traditional cutoff $z$-value for statistical significance in the Standardized
Parameter Estimates are values outside $\pm$1.96.  However, In Table 7, all of the genes
selected fail to cross this significance threshold.  In practice, users often do not know
whether the parameter estimators are normally distributed and cutoff values of 0.1 or 0.2
are often used for the standardized estimates.  In Table 7, the cutoff is 0.1 and it renders
a model with 100\% prediction accuracy.

We now turn our attention to the reliability of the neural network predictions.  To do this,
we will limit our model to a structure with only \textit{one hidden unit} to facilitate the comparison
of parameter estimates. Table 8 below shows some of the top predictors selected by the neural
network.

\begin{table}[ht!]
\caption{Top predictors selected by the neural network.}
\label{colon_journal}
\begin{tabular}{lc}
\thickline
\textbf{Gene} &	\textbf{Weight}  \\
 & \textbf{(Ranked by Absolute Value)} \\ \thickline
1769	&	-2.866915264	\\
1231	&	2.634994782	\\
1421	&	-2.243876759	\\
1702	&	-1.999461292	\\
1351	&	1.722926448	\\
258	&	-1.650588414	\\
1644	&	1.530640588	\\
1475	&	1.401072018	\\
1895	&	-1.107656933	\\
419	&	1.002626559	\\ \thickline
\end{tabular}
\end{table}

Note that in Table 8, the selection of the top genes is rather subjective and the cutoff is
arbitrary.  In practice, one can try backward elimination, forward inclusion, and the stepwise
procedure to select the genes.  However, unlike PLS and regression, the literature contains
no reliable ways to calculate the standard error of the weight and hence there is no way to
judge whether the estimates of the weights would behave like those in Table 7 (PLS) and Table
6 (regression).  The same can be said for SVM (support vector machines) and other methods in
Table 1.

In conclusion, a regression model or PLS can achieve 100\% accuracy but the parameter
estimates are not reliable.  In the current literature, the problem of ``complete separation"
of logistic regression is well-known, but there is no such analysis for PLS, neural networks,
support vector machines, or other models.  In the next section, we provide further proof that
models with high accuracy can be very misleading.

\subsection{Analysis of Prostate Cancer Data: More Miracles or More Illusions?}

We will now shift our focus from the colon cancer dataset to the benchmark prostate cancer
data that has 102 patients (52 cancer, 50 normal) and 6,033 genes.  The data was collected
and analyzed by a team of 15 scientists from a dozen institutions including Harvard Medical
School, Whitehead Institute/Massachusetts Institute of Technology, and Bristol-Myers Squibb
Inc. Princeton.

As one would imagine, it is very expensive to conduct a microarray experiment of this magnitude,
and it would be desirable to have more cost-effective alternatives.  As a result, we frame up
a scenario as follows: a biologist who has a limited budget collected only 10\% of the samples
as compared to the benchmark dataset (i.e., there are only 10 patients in the sample).  The
biologist pre-screened the 6,033 genes by a statistical variable selection technique, then ran
the PLS model with leave-one-out cross-validation, and found that the model can classify the
validation datasets as cancer or normal with 100\% prediction accuracy.  In addition, the
biologist used regression to double check the PLS results and also obtained 100\% prediction
accuracy with the same genes as the PLS model: Gene1149, Gene4201, and Gene4780. Finally, the
biologist double-checked the statistical results by examining the posterior probabilities as
shown in Table 9.

\begin{table}[ht!]
\caption{PLS posterior probabilities of the 10 patients using leave-one-out cross-validation.}
\label{table9}
\begin{tabular}{lcc}
\thickline
\textbf{Patient} &	\textbf{Mean Posterior Probability}	& \textbf{Cancer or Normal?}	\\ \thickline
1	&	1	&	Cancer	\\
2	&	1	&	Cancer	\\
3	&	0.998	&	Cancer	\\
4	&	0.995	&	Cancer	\\
5	&	0.979	&	Cancer	\\
6	&	0.011	&	Normal	\\
7	&	0.007	&	Normal	\\
8	&	0.005	&	Normal	\\
9	&	0	&	Normal	\\
10	&	0	&	Normal	\\ \thickline
\end{tabular}
\end{table}

The posterior probabilities indicate that the model did extremely well classifying the data.
This would represent an excellent finding for the biologist, especially considering that
regression and PLS use different methodologies: regression is based on the maximum likelihood
estimation of the parameters of the following equation: \vspace{-0.1in}

\begin{equation*}
 \log \left(\frac{p}{1-p}\right) = a+b_1x_1+b_2x_2+\cdot\cdot\cdot+b_kx_k+\varepsilon,
\end{equation*}

\noindent while PLS is based on the extraction of latent variables from the covariance
matrices of \vspace{-0.15in}

\begin{equation*}
 X'X  \text{ and }   Y'X.
\end{equation*}

Since the two methodologies are vastly different, the results appear to have reinforced each
other in a significant manner.  The scientist also noticed that PLS has been widely used in
analytic chemistry (see, e.g., Wold et al., 2001) and other fields (see, e.g., Vinzi et al.,
2010), so the results are very encouraging and the 10\% sample has the potential to cut
research costs by 90\%.  If the results hold water, it would be great news for all
researchers in this field of study.

But now the problem: three other imaginary scientists did the same experiment with a different
10\% of the sample.  The situation is depicted in the process flow shown in Figure 1. 

\begin{figure}[ht!]
 \begin{center}
  \includegraphics[scale=0.85]{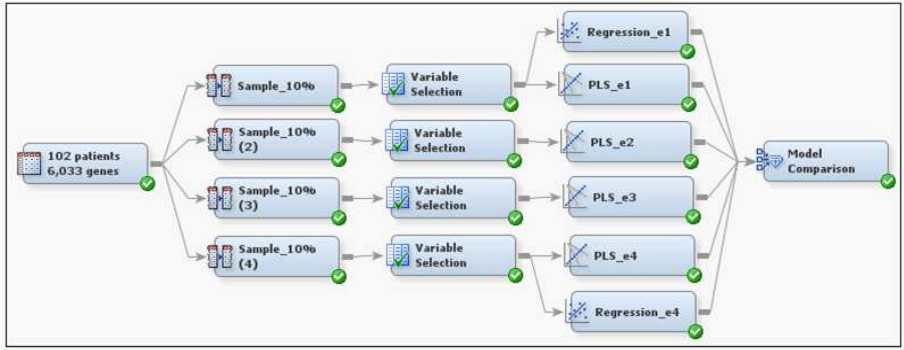}
  \caption{Process flow of the PLS and regression models with a mere 10\% sample of the prostate cancer
  data.  Here four scientists used different samples to run their models.  The first
  and the last scientists also used regression to double check their PLS results.}
  \label{Figure1}
 \end{center}
\end{figure}

\noindent Now the miracle (see Table 10 below): the four scientists all achieved 100\% prediction
accuracy but the genes they selected are vastly different.

\begin{table}[ht!]
\caption{Four different runs of PLS using leave-one-out cross validation all achieved 100\%
accuracy but the genes they selected are vastly different.  Furthermore, the genes selected
by PLS have no overlap with the genes selected by Efron’s 2010 study.}
\label{Table10}
\begin{tabular}{lclcc}
\thickline
\textbf{Method} & \textbf{Prediction} &	\textbf{Genes Selected}	& \textbf{Sample} &	\textbf{Seed} \\
   & \textbf{Accuracy} &  & \textbf{Size} &  \\ \thickline
PLS-e1	&	100\%	&	1149, 4201, 4780	&	10	&	12345	\\
PLS-e2	&	100\%	&	38, 476, 5585	&	10	&	23451	\\
PLS-e3	&	100\%	&	1352, 1751, 3560	&	10	&	34512	\\
PLS-e4	&	100\%	&	38, 1871	&	10	&	45123	\\ \hline
Efron  &	n/a 	&	610, 1720, 332, 364, 	& 102 & n/a \\
(2010) & & 914, 3940, 4546, 1068, 579, 4331 & & \\ \thickline
\end{tabular}
\end{table}

Table 10 also includes the 10 genes that were selected by Efron (2010) which have
very little in common with the rest four sets of the genes.  In summary, four scientists
set out to collect data and use PLS to find the most important genes.  In two of the four
cases, the biologists even confirmed their PLS predictions using regression.  Each of their
models has a 100\% prediction accuracy, but the genes they picked are vastly different.
Which set of genes would you believe?

We conclude that the 100\% prediction accuracy actually misled our imaginary scientists to
believe that a sample size of only 10 patients is sufficient to analyze the prostate cancer
dataset.

\section{THE RELIABILITY OF STATISTICAL METHODS (II):
RESULTS FROM SIMULATION DATA}

In this section, we will create our own data to simulate microarray data.  Comparable to
the colon cancer dataset, the simulated data will contain the expression level of 2,000
``genes" for 62 simulated ``patients".  In the colon cancer dataset, more than 15 of the
genes are correlated with a correlation coefficient $r > 0.7$ (see the scatterplot of
Gene493 and Gene249 in Figure 2). 

\begin{figure}[ht!]
 \begin{center}
  \includegraphics[scale=0.9]{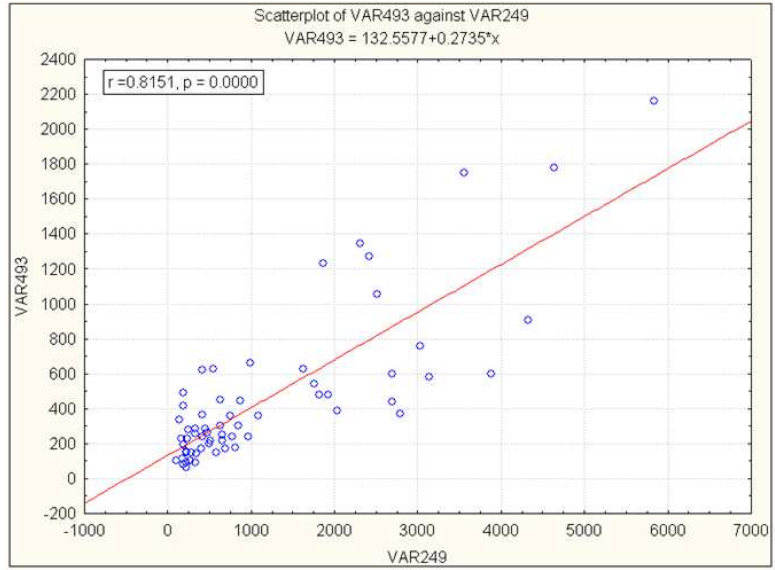}
  \caption{The scatterplot of the non-normalized expression levels of Gene493 and
  Gene249 from the colon cancer data.  The correlation of these two variables is
  0.8151 with p $\cong$ 0.0000.}
  \label{Figure2}
 \end{center}
\end{figure}

\noindent As a result, we added correlations to the three genes
{X1, X2, X3} in our simulation data.  From this point forward, Xi represents the numerical
gene expression level of gene Xi. The other 1,997 gene expression levels are generated from
a uniform distribution.

The corresponding formulae for generating correlated X1, X2 and X3 are as follows.
Assume $X_1$, $Z_1$, and $Z_2$ are independent and identically distributed random
variables.  Let \vspace{-0.15in}

\begin{equation*}
 X_2 = X_1 + bZ_1 \\
\end{equation*} \vspace{-0.45in}

\begin{equation*}
 X_3 = X_2 + cZ_2.
\end{equation*}

\noindent Using b=0.8 and c=0.75 gives the following correlations between X1, X2 and X3:

\vspace{-0.1in}

\begin{equation*}
 \rho(X_1,X_2) = \frac{\sigma^2(X_1)}{\sigma(X_1)\sqrt{\sigma^2(X_1)+b^2\sigma^2(Z)}}
 = \frac{\sigma^2(X_1)}{\sigma^2(X_1)\sqrt{1+b^2}} = \frac{1}{\sqrt{1+b^2}} \cong 0.78,
\end{equation*}

\begin{equation*}
 \rho(X_1,X_3) = \frac{1}{\sqrt{1+b^2+c^2}} \cong 0.67,
\end{equation*}


\begin{equation*}
 \rho(X_2,X_3) = \frac{\sqrt{1+b^2}}{\sqrt{1+b^2+c^2}} \cong 0.86.
\end{equation*}

The correlations of the three predictors in the simulation data are shown in Figure 3.

\begin{figure}[ht!]
 \begin{center}
  \includegraphics[scale=0.85]{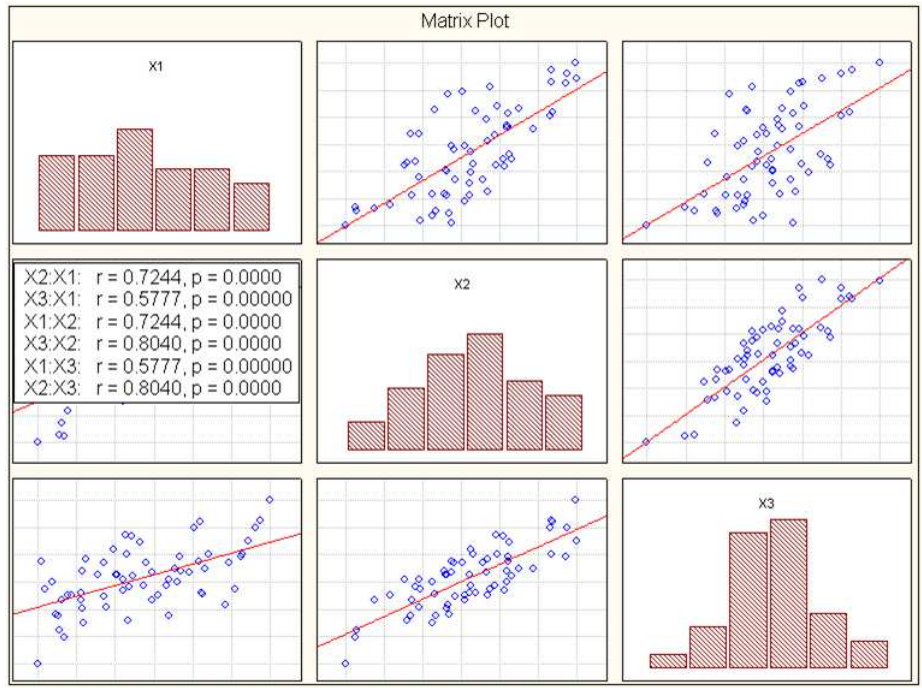}
  \caption{The scatterplots and the correlations among X1, X2, and X3.}
  \label{Figure3}
 \end{center}
\end{figure}

To facilitate the simulations of 5-gene and 10-gene interactions, we re-scaled the ranges of the variables.  The mean, standard deviation, maximum, and minimum values of {X1, X2, X3} are listed in Table 11.

\begin{table}[ht!]
\caption{Mean, standard deviation and range of genes X1, X2, X3 when n = 62.}
\label{Table 11}
\begin{tabular}{lclcc}
\thickline
Gene & Mean	& Standard Deviation & Minimum & Maximum \\ \hline
X1	&	43.9	&	26.4	&	0.4	&	97.2	\\
X2	&	166.3	&	67.2	&	23.5	&	283	\\
X3	&	278.3	&	92.4	&	35.2	&	481.3	\\
\thickline
\end{tabular}
\end{table}

We will use this controlled dataset to represent the gene expression level of 62
patients.  Then, we will define functions that classify the 62 patients as cancerous
or normal.  This will allow us to examine how the various statistical methods perform
at classifying the simulated data as diseased or normal.  The advantage of this approach
is that we already know how each dataset should be classified, and we also know which
gene expression levels are responsible for that classification of the disease.

There is precedent for using simulated data to rigorously examine the reliability of a
statistical model.  For instance, Park and Hastie (2008) investigated gene-gene and
gene-environment interactions using three discrete epistatic models and a heterogeneity
model of two interacting genes.  Each of the two genes is assumed to have a dominant
allele (form) and a recessive allele, and the models captured different potential modes
of interaction between the two genes.  Noisy data was generated from each of these
models, and statistical methods along with multifactor dimensionality reduction were used
to train and classify the simulated datasets.

What Park and Hastie did was model the interaction between the two genes of interest
in their study (2008).  However, in the case of microarray data, there are potentially
thousands of interacting genes.  This greatly complicates the analysis of microarray
data, as the scale makes it nearly impossible to model the gene-gene interactions.  To
put this in perspective, Cordell (2009, \textit{Nature Reviews Genetics}) wrote an extensive
review on detecting gene-gene interactions that underlie human disease.  The review
discussed different methods for deciphering all two-locus interactions and the
associated computational costs of each method.  The article concluded ``an exhaustive
search of all three-way, four-way or higher-level interactions seems impractical in a
genome-wide setting."  This point was driven further home in a recent article by Van
Steen entitled ``Travelling the world of gene-gene interactions" (2011,
\textit{Briefings in Bioinformatics}).  Given this reality, we cannot expect to build
models that will accurately capture the interaction between all genes that give rise
to cancer. Therefore, we cannot build a discrete, allele-based model comparable to
that of Park and Hastie for our purposes.  Instead, we will have to generate
expression-level datasets that are comparable to datasets generated from microarray
experiments. We then need mathematical equations that can classify the dataset as
diseased or normal based on the expression level of a hand-selected set of genes.

\subsection {The Simulated Diseases}

In order to classify our simulated datasets, we have designed equations that take the
dataset as input, and output whether the dataset represents a diseased state or a
normal state.  We start with a disease in which only a single gene is responsible for
the disease, and we build up from there adding more contributing genes, and more
complex (nonlinear) interactions between the genes.

Our simulated dataset consists of 2,000 genes/predictors (X1-X2000) for 62 ``patients"
and eight initial equations that will be used to classify the simulated patients as
diseased or normal.  In the first three disease equations, each gene linearly contributes
to the disease state, and there are no gene-gene interactions.

\begin{itemize}
\item
\textbf{Disease1}: disease or normal is determined solely by the expression level of X1:

\begin{equation*}
f_1 = \left\{ \begin{array}{ll}
0, & \textrm{if }X_1>53.1 \\
1, & \textrm{otherwise}
\end{array} \right.
\end{equation*}

where 0 represents a normal dataset and 1 represents a diseased dataset.  This is
similar to a number of single gene diseases, including hemophilia A (X-linked
recessive disease determined by F8 gene), cystic fibrosis (autosomal recessive disease
determined by CFTR gene), sickle-cell anemia (autosomal recessive disease determined
by HBB gene) and Huntington's disease (autosomal dominant disease determined by HTT
gene) to name a few (Chial, 2008).

\item
\textbf{Disease2}: disease or normal is determined by a linear combination of {X1, X2}:

\begin{equation*}
f_2 = \left\{ \begin{array}{ll}
0, & \textrm{if }2X_1 + X_2 > c_2 \\
1, & \textrm{otherwise}.
\end{array} \right.
\end{equation*}

This is similar to the familial breast cancer, which is attributed to two genes: BRCA1
and BRCA2 (Ritchie et al., 2001).  Note that familial breast cancer is rare (about 5\%
of the female population). For the non-familial breast cancer, the genetic structure is
a lot more complicated.

\item
\textbf{Disease3}: disease or normal is determined by a linear combination of {X1, X2, X3}:

\begin{equation*}
f_3 = \left\{ \begin{array}{ll}
0, & \textrm{if }2X_1 + 0.7X_2 + 1.5X_3> c_3 \\
1, & \textrm{otherwise}.
\end{array} \right.
\end{equation*}

Both colon cancer and prostate cancer involve more than three genes (see, e.g., Table 1
and Table 10).  Our analysis will start with three genes and gradually move upward.

\end{itemize}

The cutoff value in each function was designed so that the distribution of the disease
is \textit{relatively balanced} as shown in the colon cancer and prostate cancer data.  For instance,
the Disease 1 function $f_1$ used a cutoff of 53.1 to keep the number of cancer and normal
patients from being heavily skewed in either direction.  Figure 4 displays the histograms
of diseased and normal cases using Disease1 and Disease2.  In the remaining diseases, the
cutoffs were similarly chosen so that the distribution of disease is relatively balanced.
Instead of presenting the exact cutoff value in the other diseases, we will simply use
$c_i$ to represent the cutoff chosen for disease {\it i}.

\begin{figure}[ht!]
 \begin{center}
  \includegraphics[scale=1]{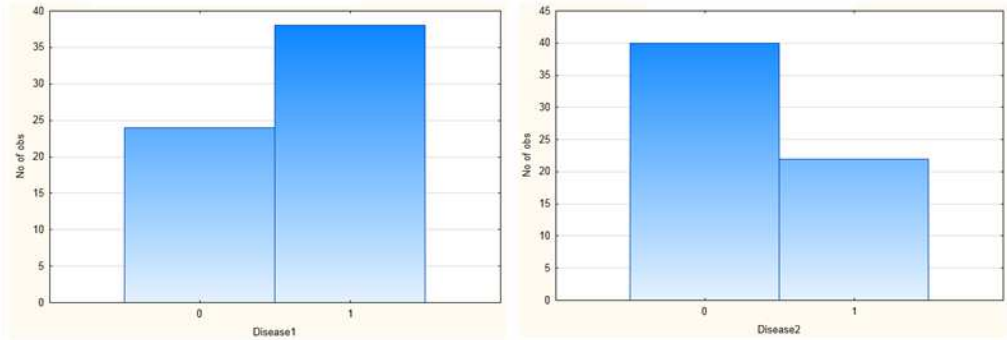}
  \caption{The distributions of the Disease1 and Disease2 are relatively balanced, with
  1 indicating cancer, and 0 indicating normal.}
  \label{FIG 4}
 \end{center}
\end{figure}

\newpage

In each of Disease1 through Disease3, only linear combinations of different genes are
considered, so no gene-gene interactions take place, and each gene contributes independently
to the disease state.  Disease4 through Disease8 each account for \textit{nonlinear} contributions of
genes, and gene-gene interactions.

\begin{itemize}
\item
\textbf{Disease4} is determined by a nonlinear combination of three genes (X1, X2 and X3)
with the nonlinear term in X1:

\begin{equation*}
f_4 = \left\{ \begin{array}{ll}
1, & \textrm{if }X_1^2 + X_2 + X_3 > c_4 \\
0, & \textrm{otherwise}.
\end{array} \right.
\end{equation*}

This represents a scenario in which X1 is a stronger determinant of the disease than X2
or X3.

\item
\textbf{Disease5} is a modification of Disease4 where X1, X2 and X3 are centered around
their means ($\mu_i$ is the mean of gene Xi):

\begin{equation*}
f_5 = \left\{ \begin{array}{ll}
1, & \textrm{if }(X_1 - \mu_1)^2 + (X_2 - \mu_2) + (X_3 - \mu_3) > c_5 \\
0, & \textrm{otherwise}.
\end{array} \right.
\end{equation*}

\item
\textbf{Disease6} is similar to Disease4, except there is a nonlinear term for both the
expression level of X1 and X2:

\begin{equation*}
f_6 = \left\{ \begin{array}{ll}
1, & \textrm{if }X_1^2 + X_2^2 + X_3 > c_6 \\
0, & \textrm{otherwise}.
\end{array} \right.
\end{equation*}

\item
\textbf{Disease7} is determined by a nonlinear combination of three genes:

\begin{equation*}
f_7 = \left\{ \begin{array}{ll}
1, & \textrm{if }X_1X_2 + X_2X_3 + X_1X_3 > c_7 \\
0, & \textrm{otherwise}.
\end{array} \right.
\end{equation*}

In this case, all gene interaction occurs in a pair-wise fashion.  The effect of
each gene individually is not considered.

\item
\textbf{Diseas8} is determined by a nonlinear combination of three genes:

\begin{equation*}
f_8 = \left\{ \begin{array}{ll}
1, & \textrm{if }X_1X_2X_3 > c_8 \\
0, & \textrm{otherwise}.
\end{array} \right.
\end{equation*}

In this case, all three genes interact and mutations in a single gene, or even two
genes, have no independent effect on the disease state.

\end{itemize}

\subsection{Machine-Learning and Predictive Modeling of Simulated Diseases}

In their ground-breaking paper in \textit{Cancer Cell}, Singh et al. (2002) used
$K$-nearest neighbor for binary classification and obtained 90\% accuracy in the
prediction of prostate cancer.  Furthermore, they maintained that (p. 206):

\begin{quote}
 The successful prediction of patient outcome will ultimately lead to improved
 decision making regarding current therapeutic options and the rational selection
 of patients at high risk for relapse for clinical trials testing adjuvant
 therapeutics.
\end{quote}

\noindent We agree with this statement.  But a cautionary note is that certain predictive
models may produce 100\% accuracy yet pick only \textit{irrelevant genes}; that is, genes
that are not implicated in the disease state (see results of Disease5 below).
We summarize our findings in Tables 12 and 13. 

\begin{table}[ht!]
\caption{Gene selections and error rates of various models when $n = 62$ patients.
Bold-faced genes indicate when the model captured all the important genes. ``$\hdots$"
indicates a set of irrelevant genes.}
\label{simulation_disease1to4}
\begin{tabular}{lllll}
\thickline
 & \textbf{Disease1} & \textbf{Disease2} & \textbf{Disease3} & \textbf{Disease4} \\ \thickline
 True Genes & X1 & X1, X2 & X1, X2, X3 & X1, X2, X3 \\
Interaction & Linear & Linear & Linear & Nonlinear \\
Type &  &  &  & \textcolor{red}{X1: most} \\
 & & & & \textcolor{red}{important} \\ \thickline
 \multicolumn{5}{c}{Genes selected by model} \\
 \multicolumn{5}{c}{(error rate)} \\ \thickline
Decision Tree & \colorbox{SkyBlue}{X1} & \colorbox{SkyBlue}{X1, X2} & X1, X2 &
\colorbox{SkyBlue}{X1}  \\
 & (0\%) &  (1.6\%) &  (4.8\%) & (0\%) \\ \hline
Boosting & \colorbox{SkyBlue}{X1}  & \colorbox{SkyBlue}{X1, X2}, $\hdots$ &
\colorbox{SkyBlue}{X1, X2, X3}, $\hdots$ & \colorbox{SkyBlue}{X1} \\
 & (0\%) & (0\%) & (0\%) & (0\%) \\ \hline
PLS & \colorbox{SkyBlue}{X1}, X666  & \colorbox{SkyBlue}{X1, X2} & X1, X3, X1009 &
\colorbox{SkyBlue}{X1}, $\hdots$ \\
 & (0\%) & (3.3\%) & (3.2\%) & (0\%) \\ \hline
NN & \colorbox{SkyBlue}{X1}, $\hdots$  & \colorbox{SkyBlue}{X1, X2}, X1025 &
X1, X3, $\hdots$ & \colorbox{SkyBlue}{X1},$\hdots$ \\
 &  (0\%) & (0\%) & (5.6\%) & (11.8\%) \\ \hline
Reg-stepwise & None  & X2  & X1, X540  & None \\
 & & (6.3\%) &  (27.8\%) &  \\ \hline
Regression default & \colorbox{SkyBlue}{X1} & \colorbox{SkyBlue}{X1, X2}, $\hdots$ &
X3, $\hdots$ & \colorbox{SkyBlue}{X1} \\
 & (0\%) & (0\%) & (5.9\%) & (0\%) \\ \thickline
\end{tabular}
\end{table}


For the linear diseases 1 and 2, each model with the exception of stepwise regression
did an excellent job classifying the data.  The statistical methods each identified
the genes that contribute to the disease state, and classify the data with at least
96\% accuracy.  While some of the statistical methods identified genes that were not
implicated in the disease state, these false discoveries are much less worrisome to
biologists than false non-discoveries.  For Disease 3, only gradient boosting selected
all of the important genes.

Disease 4 is the first nonlinear disease we examined.  In the formula to compute
Disease4, the expression level of X1 is squared, meaning it represents the \textit{most
important gene}, with X2 and X3 being less influential.  This may be the reason that
three models (decision tree, gradient boosting, and regression) picked up only X1 and
subsequently had a 0\% error rate.  While identifying all contributing genes is most
desirable, identifying the major contributing gene is most important from a biological
perspective, so these three methods can adequately handle Disease4. 

\begin{table}[ht!]
\caption{Gene selections and error rates of various models for n = 62 patients.
Bold-faced genes indicate when the model captured all the important genes. }
\label{Table 13}
\begin{tabular}{lllll}
\thickline
 & \textbf{Disease5} & \textbf{Disease6} & \textbf{Disease7} & \textbf{Disease8} \\
\thickline
True Genes	& X1, X2, X3 & X1, X2, X3 & X1, X2, X3 & X1, X2, X3 \\ \hline
Interaction & Nonlinear & No Interaction & Nonlinear & Nonlinear  \\
Type &  \textcolor{red}{X1 is the most} & \textcolor{red}{X1, X2: more} & 2-way interaction & 3-way interaction \\
 & \textcolor{red}{important}	 & \textcolor{red}{important} &  &  \\	\thickline
 \multicolumn{5}{c}{Genes selected by model} \\
 \multicolumn{5}{c}{(error rate)} \\ \thickline
Decision &	\colorbox{SkyBlue}{X1} & X2 & X2, $\hdots$ & X1, X2 \\   	
Tree  & (0\%) & (3.2\%) &  (3.2\%) &  (4.8\%)	\\  \hline

Boosting &	\colorbox{SkyBlue}{X1}, $\hdots$ &
\colorbox{SkyBlue}{X1, X2}  & X3, X10 & X1, X2, $\hdots$ \\
& (11.8\%) & (0\%) &  (11.8\%) &  (11.8\%) \\ \hline

PLS & \colorbox{yellow}{irrelevant genes} & none &	X2, X3, $\hdots$ & X1, X2 \\
 & \colorbox{yellow}{(0\%)} & & (0\%) & (9.7\%) \\ \hline
NN & \colorbox{yellow}{irrelevant genes} & $\hdots$ & X2, X3,
$\hdots$ & X1, X2, $\hdots$ \\
 & \colorbox{yellow}{(5.9\%)} & (44.4\%) &  (0\%) & (0\%) \\ \hline
Reg- & irrelevant genes  & X2  & X2, X616 & X1, X2 \\	
stepwise & (29.4\%) &  (11.1\%) & (23.5\%) & (5.9\%) \\	\hline
Reg-default & \colorbox{yellow}{irrelevant genes} &
$\hdots$ & X2, $\hdots$ & X1, X2 \\
 & \colorbox{yellow}{(0\%)} & (44.4\%) & (3.2\%) & (4.8\%) \\ \hline
LASSO & \colorbox{yellow}{irrelevant genes} &
\colorbox{SkyBlue}{X1, X2} & X2, X3, & X1, X2, \\
& \colorbox{yellow}{(0\%)} & (0\%) & $\hdots$  (0\%) & $\hdots$ (0\%) \\

\thickline
\end{tabular}
\end{table}

Turning our attention to nonlinear diseases \textit{with} gene interactions, we begin to
notice the statistical methods have difficultly identifying all contributing genes
(Table 13).  While there are several statistical methods for Disease6 through Disease8
that can identify two of the three contributing genes, there is not a single
statistical model that correctly identifies all three relevant genes for Disease7 or
Disease8.  In subsequent analysis, we will show that with an increase of sample size
from $n = 62$ to 102, Boosting can do very well with 3-gene interactions.  For
higher-order interactions (5 genes or 10 genes), we will show that larger samples are
needed.

Disease5 (first column of Table 13) gave another surprising result.  Here PLS, LASSO,
and default regression achieved 0\% error rates, but the genes they picked are purely
irrelevant. This raises a red flag: how can the statistical methods achieve 0\%
error rates when the genes that determined the disease state are not even being
considered?

Note that the results of LASSO is compatible with the findings in a forthcoming
\textit{Statistical Science} article (Huang et al., p. 14-15): LASSO ``selects 17 genes
out of 30 and 435 markers out of 532, failing to shed light on the most important
genetic markers," \url{http://www.imstat.org/sts/future_papers.html}.  Here we used
the least angle regression with 5-fold cross-validation to run LASSO.  The results are
not very encouraging.

Fortunately, both decision tree and gradient boosting were able to correctly
identify X1 as the major contributor of Disease5, though they were unable to get the
minor contributors.  In fact, in two clean cuts, decision tree faithfully picks X1
with 100\% prediction accuracy using leave-one-out cross-validation (Figure 5).  This
stands in stark comparison to other popular models like regression and neural network,
which select all irrelevant genes.  This is an indication that the tree family methods
(decision tree and boosting) may be better at detecting the underlying structure of
the gene interactions.

\begin{figure}[ht!]
 \begin{center}
  \includegraphics[scale=0.9]{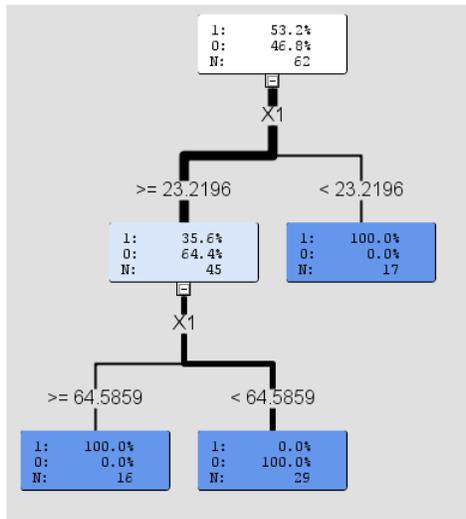}
  \caption{While popular models such as neural network, PLS, and regression selected
  irrelevant genes with 100\% prediction accuracy, the decision tree was able to pick
  the correct variable in two cuts.  The three end-nodes of the tree show
  100\% prediction accuracy in binary classification.  }
  \label{Figure5}
 \end{center}
\end{figure}

A related note is that Disease5 is not representative of any known disease in the sense
that there is no biological basis to support the centering of the variables around their
means.  The centering created the following histograms for cancer (right panel, Figure 6)
and normal patients (left panel, Figure 6):

\begin{figure}[ht!]
 \begin{center}
  \includegraphics[scale=0.9]{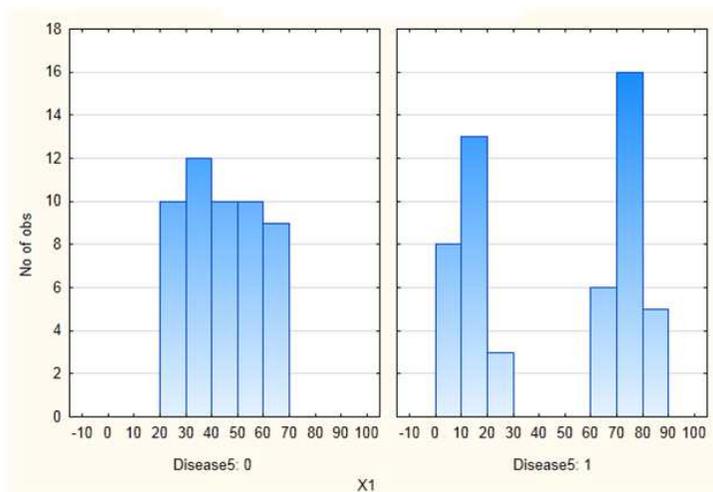}
  \caption{The right panel has a big gap in the middle of the histogram and is unlikely
  to represent any biological disease; instead, it is a ``statistical disease"
  which causes certain models to select irrelevant genes with 100\% classification accuracy. }
  \label{Figure 6}
 \end{center}
\end{figure}

In Figure 6, The histogram on the right has a big gap in the middle and is unlikely
to happen for cancer or any biological diseases; instead, it is a ``statistical disease"
which causes three popular models (neural network, PLS, and regression) to select
irrelevant variables with 100\% classification accuracy.  We believe there are many
other statistical diseases like the one presented here.  With time and effort, we may
found more examples to expose the weaknesses of statistical models that would
ultimately strengthen the science of statistical disciplines.

A final detail about Disease1 through Disease8 is that we generated our own data so
that the numbers of cancer and non-cancer patients are relatively balanced.  In reality,
this is not the case.  In some scenarios, we tried to create 1,000 patients, where 97\%
are normal and 3\% have cancer.  The data is lopsided and none of the models we tried
were able to handle it.  Consequently, we used an \textit{oversampling} technique to select all
cancer patients and an equal number of normal patients.  The oversampling process did
work well, and gave results that are similar to what we have in this Section.

\section{SAMPLE SIZE}

In Section 3, the simulation datasets were motivated by the colon cancer benchmark data
which has 62 patients and 2,000 genes.  At this sample size, there were several statistical
methods that worked well when a linear combination of genes caused the disease.  However,
the models did not work well with nonlinear interactions of three genes.  In this section,
we study the effect of both sample size and the number of contributing genes on the
reliability and accuracy of several statistical models.

\subsection{Three important genes, 102 patients}

Recall that in Section 3.2, when there are only three important genes implicated in the
simulated disease, certain models did well with linear equations (Disease1 through Disease4)
but not so for the nonlinear equations with gene-interactions ($n$ = 62).  In this section we
increase the sample size from $n$ = 62 to $n$= 102 (with $n$ = 102 being motivated by the prostate
cancer dataset), and consider Disease6 through Disease8 as defined in Section 3.  With this
moderate increase of sample size, gradient boosting \textit{picks all important genes} and its
prediction accuracies range from 93\% to 100\% as shown in Table 14 below.  In Table 14, we
also include a semi-saturated nonlinear three gene model:

\begin{equation*}
f_9 = \left\{ \begin{array}{ll}
1, & \textrm{if }X_1+X_2+X_3+X_1X_2+X_2X_3+X_3X_1+X_1X_2X_3> c_9\\
0, & \textrm{otherwise}
\end{array} \right.
\end{equation*}

\begin{table}[ht!]
\caption{Gradient boosting picks all important genes with 3-gene nonlinear relationships and
$n = 102$ patients, while decision tree does not.}
\label{Table 14}
\begin{tabular}{lcccc}
\thickline
 & \textbf{Disease5} & \textbf{Disease6} & \textbf{Disease7} & \textbf{Disease8} \\ \thickline

True 	& X1, X2, X3 & X1, X2, X3 & X1, X2, X3 & X1, X2, X3 \\
Genes	& No Interaction & 2nd order & 3rd order  & 3rd order 	\\
&  \textcolor{red}{X1, X2: more} & interaction & interaction & interaction 	\\
& \textcolor{red}{important} & & & semi-saturated \\ 	\thickline
 \multicolumn{5}{c}{Genes selected by model} \\
 \multicolumn{5}{c}{(error rate)} \\ \thickline
Decision &	\colorbox{SkyBlue}{X1, X2} & X2, X3 & X2 & X2 \\
Tree & (7.1\%) & (2.5\%) & (2.9\%) & (3\%) \\ \hline
Gradient & \colorbox{SkyBlue}{X1, X2, X3} & \colorbox{SkyBlue}{X1, X2, X3} &
\colorbox{SkyBlue}{X1, X2, X3} & \colorbox{SkyBlue}{X1, X2, X3} \\
Boosting & (5.9\%) &  (6.7\%) & (6.7\%)& (6.7\%) \\ \thickline
\end{tabular}
\end{table}

Recall that gradient boosting did not perform well on Disease7 and Disease8 when $n = 62$
patients (Table 13).  Table 14 above shows that if $n$ = 102 patients, then the model picks
all the important genes with high prediction accuracy, and it even gets the relevant minor
contributing genes (see Disease6 in Table 14).

\subsection{Five important genes, 102 or 204 patients}

In this section we go further to include 5 important genes in the simulation study:

\begin{equation*}
\a
f_{10} = \left\{ \begin{array}{ll}
1, & \textrm{if }\Pi_{i=1}^5X_i > c_{10} \\
0, & \textrm{otherwise}
\end{array} \right.
\end{equation*}

\noindent Table 15 below summarizes the results with gradient boosting and the Benjamini-Hochberg
Fdr procedure.  We found that the results from adjusted $p$-values by Fdr (Benjamini and
Hochberg, 1995) and adaptive Fdr (Benjamini et al., 2006) are compatible to one another.

\begin{table}[ht!]
\caption{Diease10, which depends on the five genes X1-X5, using $n = 102$.
Benjamini-Hochberg Fdr picked only four genes unless the number of genes is prescreened
to 100.  Gradient boosting, in comparison, was able to pick all five genes from a
prescreened pool of 2,000 genes.}
\label{Table 15}
\begin{tabular}{lll}
\thickline
\textbf{Method}	& $\boldsymbol{n = 102}$ \textbf{patients} & $\boldsymbol{n = 204}$
\textbf{patients} \\ \thickline
Boosting-1 & X1-X4 & X1-X5  \\
(6,033 genes) & (missed X5) & accuracy = 86.3\% \\ \hline
Boosting-2 & X1-X5 & X1-X5 \\
(2,000 genes) & Accuracy = 79\% & accuracy = 86.3\%  \\ \thickline
Fdr-1 &	X1-X4 &	X1-X5 \\
(6,300 genes) & (missed X5) & + 2 other genes \\ \hline
Fdr-2 &	X1-X4 & X1-X5 \\
(2,000 genes) & (missed X5) & + 6 other genes \\ \hline
Fdr-3  & X1-X5 & X1-X5 	\\
(100 genes)  &	 +1 other gene &  \\ \thickline
\end{tabular}
\end{table}

Table 15 shows that if $n = 204$, then both gradient boosting and the Benjamini-Hochberg
Fdr procedure would be able to pick up all important genes.  Note that microarray
experiments are very expensive (although the price is decreasing in recent years), and a
large sample like $n = 204$ may be beyond the reach of many scientists.  Consequently a
procedure that can handle small sample is highly desired.

Turning to the case when $n = 102$ patients, both gradient boosting and Fdr missed one
important gene, which is \textit{not acceptable} to biologists -  once an important gene is lost,
then it cannot be recovered.   Nevertheless, if the number of the genes is cut from
6,033 to 2,000 using a pre-screening method, then boosting would succeed, but this is
not the case with Fdr.  The problem with Fdr persisted when we cut the number of genes to
500 (not shown in Table 16).   But if the number of genes is cut to 100, then Fdr would
succeed.

Gradient boosting is well-known for being able to model nonlinear phenomena (Friedman,
2001, \textit{Annals of Statistics}), but if there are too many genes in the model (e.g.,
6,033 genes, too much noise) and $n$ is relatively small (e.g., 102 patients), then the
model would fail.  For this reason, in Table 15, we use gradient boosting to rank the
predictors, cut the bottom ones and then re-fit the model.  This procedure may raise
eyebrows if we do it with Fdr (where $p$-values are involved) and one may argue that
the repeated adjustments of $p$-values would violate the validity of statistical inference.
In our opinion, both gradient boosting and Fdr are \textit{exploratory} tools in the gene
selection, and hence the issue of statistical inference does not really matter.

\subsection{Ten important genes, $n=$ 102, 204, 306, or 408 patients}

We now extend our simulation to include 10 important genes:

\begin{equation*}
f_{11} = \left\{ \begin{array}{ll}
1, & \textrm{if } \a \Pi_{i=1}^{10}X_i > c_{11} \\
0, & \textrm{otherwise}
\end{array} \right.
\end{equation*}

\noindent The results are shown in Table 16 below.

\begin{table}[ht!]
\caption{Ten important genes causing Disease 11.  In each cell, we give the number of
important genes selected by the specified statistical method, and ``$\hdots$" indicates a
set of irrelevant genes were also chosen.  We only give data for False Discovery and
accuracy when the statistical method succeeded at finding all 10 relevant genes.  We
do this because FNDs (False Non-Discoveries) are not acceptable: once an important gene
is lost, then it cannot be recovered.  Note that PLS is not consistent.}
\label{Table 16}
\begin{tabular}{lllll}
\thickline
 & $\boldsymbol{n = 102}$  & $\boldsymbol{n = 204}$ & $\boldsymbol{n = 306}$ &
 $\boldsymbol{n = 408}$ \\ \thickline
Boosting-1 & 1 gene  & 6 genes  & 8 genes   & 9 genes  \\
(6,033 genes) & $+\hdots$ & $+\hdots$ & $+\hdots$ & $+\hdots$ \\ \hline

Boosting-2 & 4 genes & 9 genes & \colorbox{SkyBlue}{10 genes} & \colorbox{SkyBlue}{10 genes} \\
(500 genes) &  $+\hdots$ & $+\hdots$  & FDiscovery = 98\% & FDiscovery = 98\%  \\
 &  &  & accuracy = 83\% & accuracy = 87\% \\ \hline

Boosting-3  & 9 genes & \colorbox{SkyBlue}{10 genes} & \colorbox{SkyBlue}{10 genes}  &
\colorbox{SkyBlue}{10 genes} \\
(100 genes) & $+\hdots$  & FDiscovery = 90\% & FDiscovery = 90\% & FDiscovery = 90\%  \\
 & & accuracy = 80\% & accuracy = 86\% & accuracy = 88\% \\ \hline

Boosting-4 	& 9 genes  &  \colorbox{SkyBlue}{10 genes} & \colorbox{SkyBlue}{10 genes} &
\colorbox{SkyBlue}{10 genes} \\
(20 genes) & $+\hdots$  & FDiscovery = 20\% & FDiscovery = 20\% & FDiscovery = 20\% \\
& & accuracy = 87\% & accuracy = 91\% & accuracy = 88\% \\ \hline

Boosting-5 & 9 genes & \colorbox{SkyBlue}{10 genes} & \colorbox{SkyBlue}{10 genes} &
\colorbox{SkyBlue}{10 genes}  \\
(12 genes) & $+\hdots$ & FDiscovery = 17\% & FDiscovery = 17\% & FDiscovery = 17\% \\
 & & accuracy = 88\% & accuracy = 92.2\% & accuracy = 94.5\% \\ \thickline

Fdr & 1 gene  & 5 genes & 8 genes &	7 genes \\
(6,033 genes) & + X1379 & & + X1502 & \\ \hline
Fdr & 4 genes & 5 genes & 7 genes & 8 genes \\
(2,000 genes) & & & & \\ \hline
Fdr & 4 genes & 8 genes & 9 genes & \colorbox{SkyBlue}{10 genes} \\
(500 genes) & + X37 & + X382 & 	& \\ \hline
Fdr	& 4 genes & 8 genes  & \colorbox{SkyBlue}{10 genes} & \colorbox{SkyBlue}{10 genes} \\
(100 genes) & + X59 & + X59 & & \\ \hline
Fdr & 5 genes &	\colorbox{SkyBlue}{10 genes} & \colorbox{SkyBlue}{10 genes} & \colorbox{SkyBlue}{10 genes} \\
(20 genes) & &	& & + X20 \\

\thickline

PLS-m1 & 9 genes & 9 genes & 9 genes &
\colorbox{SkyBlue}{10 genes} \\
(40 genes) & & & & FDiscovery = 0\% \\
 &  & & & accuracy = 89\% \\ \hline

PLS-m2 & 9 genes & \colorbox{SkyBlue}{10 genes}
& \colorbox{SkyBlue}{10 genes} & 9 genes \\
(20 genes) & & FDiscovery =0\% & FDiscovery = 0\% & \\
 & & accuracy = 89\%	& accuracy = 89\% & \\ \thickline

\end{tabular}
\end{table}

The data in Table 16 allows us to readily see the effects of sample size on gene selection,
accuracy and the false discovery rates of various statistical methods.  When looking at ten
interacting genes, which can reasonably be expected in cancer, a sample size of $n = 102$ is
too small for Fdr, gradient boosting and PLS, as all important genes are not identified.  These
false non-discoveries are costly to biologists, as once a gene is screened out of a pool, there
is no chance of identifying that gene as an essential component of a disease.

Increasing the sample size to $n = 204$ starts to paint a different picture.  All three methods
perform better at this sample size. However, in each case, a prescreening method is required
to cut down the number of genes used in the model.  In the case of Fdr and PLS, the prescreened
pool of genes must be of size 20 in order to identify all the relevant genes.  When the sample
size is $n=204$ patients, gradient boosting performs consistently well on prescreened pools of
size 100 genes or smaller.

As we increase the sample size, we find that Fdr’s performance improves, with all ten relevant
genes for Disease11 being selected from larger prescreened pools.  PLS's performance also improves
with sample size, but does not pick up all 10 genes when $n$ = 408 patients and there are 20 genes
in the model.  Gradient boosting is a consistent performer at larger sample sizes, provided a
prescreening procedure is applied to cut down the pool of 6,033 genes.

However, gradient boosting is well-known for being an algorithm for \textit{greedy} function approximation
(Friedman, 2001).  As a result, the false discovery rate is relatively high.  For instance, in
Table 16, Boosting-3 with $n = 204$ has an extremely high false discovery rate of 90\%.  While
it is desirable that boosting identified all 10 relevant genes, this 90\% false discovery rate
means the boosting algorithm is technically selecting all 100 prescreened genes.  This seems to
suggest that boosting is a weak model.  However Figure 7 shows that the Variable Importance
scores of the top relevant genes are much higher than the majority of the other genes, with the
exception of exactly one false discovery in the 11 most important genes.  Therefore, while
boosting is ranking all genes as important, it is ranking the top ten genes (the ten relevant
genes that determine the disease state) as significantly more important than essentially all of
the other genes.  This chart can thus be used to allow a biologist to cut down the number of genes
without resulting in the emergence of false non-discoveries.  This is precisely how we cut down
the size of the prescreened pool of genes in Boosting-4 and Boosting-5 of Table 16.  \newpage

\begin{figure}[ht!]
 \begin{center}
  \includegraphics[scale=0.8]{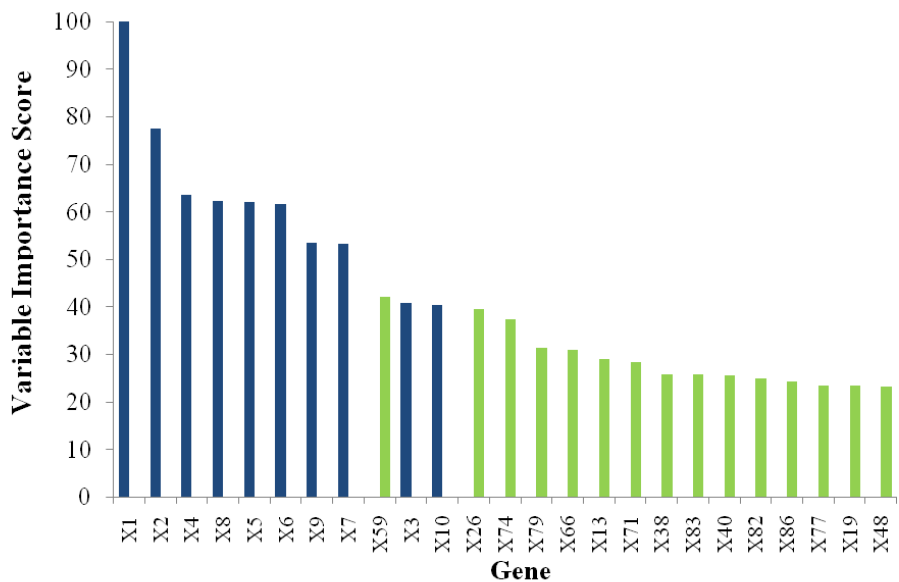}
  \caption{Gene score for the top 25 genes for Boosting-3 (Table 16) with $n = 204$ patients.
  The genes implicated in the disease are shown in dark blue and those not implicated in the
  disease are shown in light green.  In spite of the high Fdiscovery rate of 90\%, the eleven genes
  with the highest score contain all ten genes that cause the disease.}
  \label{ Figure7}
 \end{center}
\end{figure}

\subsection{Non-Taylor interactions}

In the literature, the gene-gene interactions have been modeled using Taylor polynomials (see,
e.g., Park and Hastie, 2008; Assimes et al., 2008; Cordell, 2009).  Our results showed that when gene
interactions are described by Taylor polynomials, gradient boosting is a reliable method when
coupled with prescreening and a sufficiently large sample size.  In this section, we explore a
number of non-Taylor interactions with \textit{multiple thresholds} and determine if the statistical
methods have the same level of success.  The diseases explored are:

\begin{equation*}
f_{101} = \left\{ \begin{array}{ll}
1, & \textrm{if }X_1 > 27 \text{ and } X_2 > 70 \text{ and } X_3 < 220 \\
0, & \textrm{otherwise}
\end{array} \right.
\end{equation*}

\begin{equation*}
f_{102} = \left\{ \begin{array}{ll}
1, & \textrm{if } X_1 > 23 \text{ and } X_2 > 34 \text{ and } X_3 < 180 \\
0, & \textrm{otherwise}
\end{array} \right.
\end{equation*}

\begin{equation*}
f_{103} = \left\{ \begin{array}{ll}
1, & \textrm{if } X_1X_2 > 300 \text{ and } X_3 < 140 \\
0, & \textrm{otherwise}
\end{array} \right.
\end{equation*}

For Disease101, the thresholds were chosen so that X1 and X2 are more important than X3.
For Disease102, the three genes (X1, X2, X3) have equal weights. For Disease103, X1X2 and X3
have equal weights.  Table 17 below shows the results of Fdr, gradient boosting, and PLS
models for these three diseases.

\begin{table}[ht!]
\caption{Non-Taylor interactions with three genes, X1-X3 with $n = 102$ patients. Fdr
fails on Disease102 and 103.  PLS was shown to be inconsistent in Table 16, therefore
we do not present further results for PLS.  Gradient boosting appears to be a better
tool.}
\label{Table 17}
\begin{tabular}{llll}
\thickline
 & \textbf{Disease101} & \textbf{Disease102}& \textbf{Disease103} \\
 & X1 $>$ X2 $\gg$ X3 &  X1, X2, X3 have & X1*X2 and X3 have  \\
 &  \textcolor{red}{X3 is minor} & equal weights & equal weights \\ \thickline

Fdr  & \colorbox{SkyBlue}{X1, X2}, X3 & X1, X3 & X2, X3 \\
(100 genes) & & & \\

\thickline

Boosting-1 & \colorbox{SkyBlue}{X1, X2}, $\hdots$ & \colorbox{SkyBlue}{X1, X2, X3},
$\hdots$ & \colorbox{SkyBlue}{X1, X2, X3}, $\hdots$ \\
(100 genes) & FDiscovery = 78\%  & FDiscovery = 60\% & FDiscovery = 96\% \\
 & accuracy = 87\% & accuracy = 93\% & accuracy = 100\% \\ \hline

Boosting-2 & \colorbox{SkyBlue}{X1, X2}, $\hdots$ & \colorbox{SkyBlue}{X1, X2, X3}, $\hdots$ &	
\colorbox{SkyBlue}{X1, X2, X3}, $\hdots$ \\
(20 genes)& FDiscovery = 90\% & FDiscovery = 70\% & FDiscovery = 85\% \\
& accuracy = 87\% &accuracy = 87\% &	accuracy = 93\% \\ \hline

Boosting-3 & \colorbox{SkyBlue}{X1, X2}, $\hdots$ & \colorbox{SkyBlue}{X1, X2, X3}, $\hdots$ &
\colorbox{SkyBlue}{X1, X2, X3}, $\hdots$ \\
(10 genes) & FDiscovery = 80\%  & FDiscovery = 70\% & FDiscovery = 70\% \\
 & accuracy = 87\% & accuracy = 93\% & accuracy = 93\% \\ \hline

Boosting-4  & \colorbox{SkyBlue}{X1, X2}, X3, $\hdots$ & \colorbox{SkyBlue}{X1, X2, X3}, $\hdots$ &
\colorbox{SkyBlue}{X1, X2, X3}, $\hdots$ \\
(5 genes) & FDiscovery = 40\% & FDiscovery = 40\% & FDiscovery = 40\% \\
 & accuracy = 87\% & accuracy = 93\% & accuracy = 93\% \\ \hline

Boosting-5  &	\colorbox{SkyBlue}{X1, X2}, X3 & \colorbox{SkyBlue}{X1, X2, X3}  &
\colorbox{SkyBlue}{X1, X2, X3}  \\
(3 genes) & FDiscovery = 0\%  &	FDiscovery = 0\% &  FDiscovery = 0\%   \\
 & accuracy = 87\% & accuracy = 93\% & accuracy = 93\% \\

\thickline

PLS & \colorbox{SkyBlue}{X1, X2}, X3  & X1, X2 & X2, X3 \\
(100 genes) & FDiscovery = 0\%  & &  \\
 &  accuracy = 90\% &  &  \\ \thickline

\end{tabular}
\end{table}

From Table 17, we can see that for non-Taylor gene interactions, PLS and Fdr do not produce
reliable results at a sample size of $n = 102$.  However, gradient boosting continues to prove
to be a better tool, as it can well identify the relevant genes and classify the data at a
sample size of 102 when the gene interactions are non-Taylor.

\section{SUPPORT VECTOR MACHINE}

In Table 1 (colon cancer data), we presented certain classification results from the SVM
community.  Collectively, the results in Table 1 indicate that there is room for improvement.
In this Section, we will discuss our evaluation of the SVM technology.

Note that given a set of data that is completely separated by a specific threshold such as
Disease1 through Disease11 in our simulations, \textit{theoretically} it is possible to construct two
convex hulls to separate the data (Figure 8), \newpage

\begin{figure}[ht!]
 \begin{center}
  \includegraphics[scale=1]{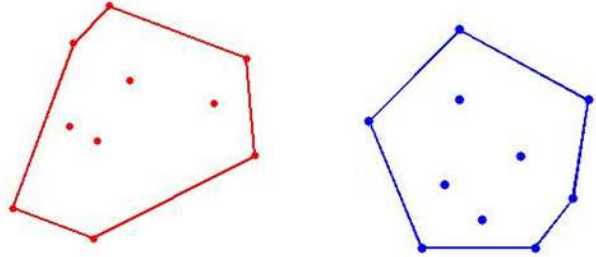}
  \caption{Convex hulls of two data sets that are completely separated from each other.}
  \label{Figure 8}
 \end{center}
\end{figure}

\noindent where a convex hull is defined as

\begin{equation*}
 C = {\{\sum_{j=1}^n {\lambda_j\overrightarrow{p_j}}:\lambda_j \geq 0, \forall j, \sum_{j=1}^n{\lambda_j=1}\}}.
\end{equation*}

Given the 2 convex hulls, one can define \textit{support vectors} to find optimal separation of the
data as shown in Figure 9.

\begin{figure}[ht!]
 \begin{center}
  \includegraphics[scale=1]{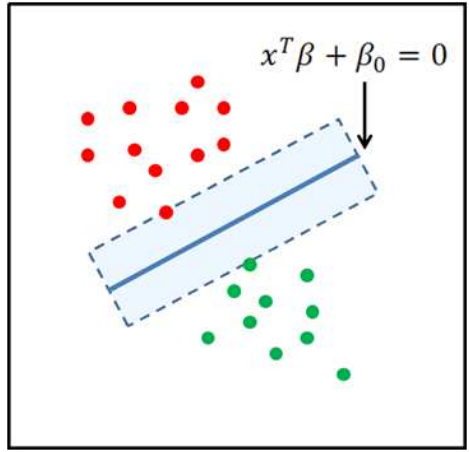}
  \caption{Optimal separation of two datasets}
  \label{ Figure9}
 \end{center}
\end{figure}

\noindent For nonlinear problems such as Disease10 and Disease 11, the goal of a support vector
machine is to transform the data from the low-dimensional space to a high-dimensional
space and then use a hyper-plane to separate the data (see, e.g., Hastie, Tibshirani,
Friedman, 2011).   When there are 6,033 genes, the prediction space is not really ``low
dimensional," but a forward-selection process starting with a single gene is feasible.
Furthermore, if the biological interactions of certain genes are roughly known, then that
piece of knowledge may guide the statistician to pick a kernel for optimal separation.

In our simulation study, Disease10 uses 5 important genes to create the target disease;
consequently we would expect the SVM technology to achieve 100\% prediction accuracy when
all 5 important genes are fed into the model.  Table 18 below shows that this is not the
case for the four different kernels we tried.

\begin{table}[ht!]
\caption{Prediction accuracy of four SVM models.}
\label{Table 18}
\begin{tabular}{lcccc}
\thickline
\multicolumn{5}{c}{Disease10 with 5 important genes: X1-X5} \\
\multicolumn{5}{c}{Accuracy of SVM  (10-fold cross-validation)} \\ \thickline
\textbf{Kernel}	& $\boldsymbol{n = 102}$ & $\boldsymbol{n = 204}$ & $\boldsymbol{n = 306}$ &
$\boldsymbol{n = 408}$ \\ \thickline
Linear	&	89\%	&	94\%	&	93\% 	&	90\% \\
Polynomial & 85\% & 92\% & 96\% &	90\%	\\
RBF	&	89\% &	95\% &	97\%	&	94\%	\\
Sigmoid	&	89\%	&	90\%	&	93\%	&	89\%	\\
\thickline
\end{tabular}
\end{table}

Table 19 shows the SVM prediction accuracy for Disease11 with 10 important genes. The
results are equally disheartening.

\begin{table}[ht!]
\caption{Prediction accuracy of four SVM models.}
\label{Table 19}
\begin{tabular}{lcccc}
\thickline
\multicolumn{5}{c}{Disease11 with 10 important genes: X1-X10} \\
\multicolumn{5}{c}{Accuracy of SVM  (10-fold cross-validation)} \\ \thickline
\textbf{Kernel}	& $\boldsymbol{n = 102}$ & $\boldsymbol{n = 204}$ & $\boldsymbol{n = 306}$ &
$\boldsymbol{n = 408}$ \\ \thickline
Linear	&	77\%	&	90\%	&	88\%	&	82\%	\\
Polynomial	&	81\%	&	86\%	&	86\%	&	84\%	\\
RBF	&	73\%	&	94\%	&	90\%	&	86\%	\\
Sigmoid	&	73\%	&	92\%	&	87\%	&	83\%	\\
\thickline
\end{tabular}
\end{table}

\noindent Table 19 also shows that the increase of sample size from $n = 204$ to $n = 306$
or 408 only confounds the model and the result is a decrease of classification accuracy.

In the SVM literature, there exists a great variety of kernel functions that can be used to
transform nonlinear to linear data.  A list of 25 kernels can be found at
(\url{http://crsouza.blogspot.com/2010/03/kernel-functions-for-machine-learning.html}).  The
kernels include Laplacian kernel, ANOVA kernel, spline kernel, Bessel kernel, Cauchy kernel,
chi-square kernel, histogram intersection kernel, generalized $t$-student kernel, Bayesian kernel,
wavelet kernel, etc.  With modifications and hybridizations, one may be able to generate hundreds
or thousands of other kernels as partially shown in Table 1 of this paper.

It would be interesting to know whether any of new kernels could actually separate the data in
Disease1 through Disease11, Disease101 through Disease103, and possibly in other cases where
gene interactions are nonlinear or non-Taylor.  We do not see an easy way to meet this challenge.
But the effort should be worthwhile and may generate new insights into this important field of
statistical learning.

The same may be true for thousands of new techniques in the field of variable selection.  In
modern statistics, the literature on variable selection is vast, complicated, and chaotic. A
systematic approach to evaluate these new tools can be a huge challenge, but it may be able to
provide scientists certain guidance on how to select tools in the important task of variable
selection.

\section{Discussions and Concluding Remarks}

In a 2008 lecture, Stanley Young of NISS (the National Institute of Statistical Sciences)
maintained that

\begin{quote}
 Empirical evidence is that 80-90\% of the claims made by epidemiologists are false; these
 claims do not replicate when retested under rigorous conditions.
\end{quote}

\noindent Young's conclusion was based on the following: findings from top medical research journals,
a survey of journal editors from diverse fields of science, and a finding that of 20 claims coming from
observational studies, only one replicated when tested in NIH funded randomized clinical trials.

In a follow-up article, Young and Karr (2011) further examined 52 claims from {\it Journal of
the American Medical Association, the New England Journal of Medicine, Journal of the National
Cancer Institute}, and {\it Archives of Internal Medicine}.  The conclusion is that ``Any claim
coming from an observational study is most likely to be wrong."

To the insiders of statistical analysis of cause-and-effect, a lot of false claims may be
avoidable if researchers take note of a motto from Rubin and Holland (see Holland, 1986):

\begin{quote}
\center \large{\textbf{No Causation Without Manipulation.}}
\end{quote} \vspace{0.1in}

\noindent The motto, of course, has exceptions even in observational studies.  For example,
in astronomy we cannot manipulate any of the relevant quantities, yet predictions in astronomy
are often more accurate than those produced by double-blind randomized controlled experiments.

In the field of gene identification, data from microarray experiments and from similar settings
are in the category of observational studies, although they are called ``experiments" in the
broad scientific community.  The use of statistical analysis to determine which gene (or set
of genes) is the cause of a specific disease is often confounded by the following factors:

\begin{enumerate}
\item
We cannot flip a coin and then assign a patient to have cancer or no cancer.  Consequently the
$t$-tests, $p$-values, Bonferrani adjustment, and Benjamini-Hochberg Fdr all lose their footing
in the statistical analysis of cause-and-effect.  Another problem with $p$-values is that
``statistical significance" is not the same as ``practical significance" and the $p$-values can
be very misleading in the gene selection process.   Furthermore, our work has shown that the
$t$-test is not efficient in the detection of non-Taylor interactions.

\item
The situation is worse with the regression model or other machine-learning tools such as neural
networks and support vector machines, even if the investigators use various randomization
techniques on the laboratory animals.  Freedman (2008, \textit{Statistical Science}) pointed out
that ``Randomization Does Not Justify Logistic Regression."  In addition, he questioned the
scientific ground of using logit versus probit or other types of the link functions in binary
regression.  Furthermore, in binary regression, often there are \textit{multiple} variables on the
right-hand side of the equation, and we simply cannot expect biologists to randomize or
manipulate the variables in the experiment.  The same can be said for other models such as
partial least squares, support vector classifier, and gradient boosting.

\item
Another problem with the statistical models and the adjusted $p$-values is that often a single gene
is responsible for the onset of a specific disease.  But when that gene is altered, it may change
the expression level of dozens of other downstream genes.  Imagine a gene that is
solely responsible for causing a specific disease; in its active state, the gene releases proteins
and then the expression of 10 or 20 other genes is affected in the process.  Now how do we expect
the shuffling of statistical methods to identify the primary gene and not pick up all the secondary
genes instead?

\end{enumerate}

Fortunately, in the area of gene identification, new techniques have been developed that would
confirm to the motto of Rubin and Holland.  As one example, biologists can create knockout organisms
in which the function of a particular gene is shut down.  Using these knockouts, a biologist is able
to infer the impact of the shut-down gene by studying how organisms with the gene differ from
organisms without the gene.  For instance, Fong and colleagues developed knockout mice for the
fragile histidine triad (FHIT) gene.  They discovered that mice without this gene are more
susceptible to carcinogen-induced tumor formation than those mice that express the gene (Fong et
al., 2000).

While experiments like this are incredibly powerful, this approach is not a feasible way to
identify unknown/undetermined genes that cause a particular disease - there are simply too many
genes in the genome for a biologist to knock them all out and observe their effect.  Statistical
analysis of microarray data reduces this large pool of genes to a reasonably-sized pool that
biologists could then examine using more rigorous experimental approaches.  Therefore, statistical
analysis of microarray data can be viewed as an important first step in identifying genes that
potentially cause a particular disease.

In this study, we found that the technique of stochastic gradient boosting (Freeman, 2001) was
able to identify all important genes from a pool of 6,033 with the sample size of $n = 102$.  We
did this with simulation datasets that involve 5-gene interactions, 10-gene interactions and
non-Taylor 3-gene interactions.  The simulations have been crafted to match the real situation as
closely as possible (Section 3) - the equations are deterministic but they also mimic the colon
cancer data and the prostate cancer data (Efron, 2010, 2008): random elements, correlations among
the genes, etc.  In all cases, the gradient boosting was able to pick the contributing genes.

In comparison, the following techniques missed important genes in various scenarios: Bonferrani
adjustment, Benjamini-Hochberg Fdr, logistic regression, partial least squares, LASSO (least angle regression), neural network, decision tree, and support vector machine.  This is problematic and points to the crucial
distinction between \textit{false discovery} and \textit{false non-discovery} in the statistical gene search,
where false discovery leads to the selection of irrelevant genes and false non-discovery misses
out important genes that cannot be recovered in the subsequent analysis.  From the biological
view point, false non-discovery is \textit{not acceptable} for the very reason that if an important gene
is lost in the statistical exploration, then it will mislead subsequent research efforts.

In addition, our investigation shows that commonly used measures in binary classifications can
be very misleading in gene identification: error rate, false positive, false negative, and other
measures that are derived from these values (sensitivity, specificity, ROC curves, the area under
the ROC curve, $F$-measures, precision, recall, etc.).  The most troubling is that some commonly
used models would produce 100\% accuracy measures and select different sets of genes.  They simply
cannot stand the scrutiny of parameter estimates and model stability.

Currently there are thousands of tools for variable selection, with new ones showing up at an
exponential rate.  The growth of this field will provide us new techniques to tackle many hard
problems with high-dimensional data.  Nevertheless, the growth also creates a problem for scientists
who are facing thousands of variables and thousands of tools to select the relevant variables.
In most cases, nobody knows which variable is causing what and existing subject knowledge often
conflicts with each other.  In many cases, the search process is like trying to find a black cat
in a dark house.

In our investigation, we compared the results from real-world data and from simulation studies.
The use of simulation is a standard practice in statistics, even college students know that Ulam
and von Newmann did it in the Manhattan Project some 70 years ago.  But in the fields of gene
search and variable selection, the literature is very shy on this technology.  Here you play god,
create the genes of your liking, investigate the sample size needed, compare the tools, and finally
select the top variables for the specific phenomenon under the study.  In our case, we found that
certain widely used models (neural network, PLS, logistic regression, and LASSO) would render 100\% prediction
accuracy with genes that are \textit{not responsible} for our simulated diseases.  On the other
hand, with moderate sample size, gradient boosting will be shown to be a superior model for gene
selection, though we suspect there are more tools that are appropriate for gene search.  We
believe a platform would be beneficial in helping to select the top tools before we try to select
the top variables.


\begin{thebibliography}{99}
\bibliographystyle{imsart-nameyear}

\bibitem{alon99}  ALON, U., BARKAI, N., NOTTERMAN, D. A., GISH, K., YBARRA, S., MACK, D., and  LEVINE, A. J. (1999). Broad patterns of gene expression revealed by clustering analysis of tumor and normal colon tissues probed by oligonucleotide arrays. {\it Proc. Natl. Acad. Sci.} \textbf{96} 6745–-6750.

 \bibitem{2} ASSIMES T. L., KNOWLES, J. W., BASU, A., IRIBARREN, C., SOUTHWICK, A., et al. (2008).  Susceptibility locus for clinical and subclinical coronary artery disease at chromosome 9p21 in the multi-ethnic advance study.  {\it Hum Mol Genet.} \textbf{17} 2320--2328.

 \bibitem{2} BENJAMINI, Y. and HOCHBERG, Y. (1995). Controlling the false discovery rate: a practical and powerful approach to multiple testing. {\it Journal of the Royal Statistical Society, B}. \textbf{57} 289–-300.

 \bibitem{2} BENJAMINI, Y., KRIEGER, A. M., and YEKUTIELI, D. (2006). Adaptive linear step-up false discovery rate controlling procedures. {\it Biometrika}. \textbf{93} 491-–507.

 \bibitem{2} BAR, H., BOOTH, J., SCHIFANO, E., and WELLS, M. T. (2010). Laplace approximated EM microarray, analysis: an empirical Bayes approach for comparative microarray experiments. {\it Statistical Science}. \textbf{25} 388-–407.

 \bibitem{2} BLACK, M. A. and DOERGE, R. W. (2002). Calculation of the minimum number of replicate spots required for detection of significant gene expression fold change in microarray experiments. {\it Bioinformatics}. \textbf{18} 1609–-1616.

\bibitem{2}BREIMAN, L., FRIEDMAN, J. H., OLSHEN, R. A., STONE, C. J. (1983). {\it Classification and Regression Trees}. Chapman and Hall/CRC.

\bibitem{2}CHIAL, H. (2008). Rare genetic disorders: Learning about genetic disease through gene mapping, SNPs, and microarray data. {\it Nature Education}.

\bibitem{2}CORDELL, H. J. (2002). Epistasis: what it means, what it doesn’t mean, and statistical methods to detect it in humans. {\it Human Molecular Genetics}. \textbf{11} 2463–-2468.

\bibitem{2}CORDELL, H. J. (2009). Detecting gene–gene interactions that underlie human diseases. {\it Nature Reviews Genetics}. \textbf{ 10} 392--404.

\bibitem{2}DEAN, N. and RAFTERY, A. E. (2010). Latent class analysis variable selection.  {\it Ann Inst Stat Math}. \textbf{62} 11-–35.

\bibitem{2} DUDOIT, S., SHAFFER, J. P., and BOLDRICK, J. C. (2003). Multiple hypothesis testing in microarray experiments. {\it Statistical Science}. \textbf{18} 71–-103.

\bibitem{2} EFRON, B. (2008). Microarrays, empirical Bayes and the two-groups model. {\it Statistical Science}. \textbf{23} 1-–22.

\bibitem{2} EFRON, B. (2010). The future of indirect evidence. {\it Statistical Science}. \textbf{25} 145-–157.

\bibitem{2} EFRON, B. and Zhang, N. (2011). false discovery rates and copy number variation. {\it Biometrika}.  \textbf{98} 251--271.

\bibitem{2} FERREIRA, J. A. and ZWINDERMAN, A. H. (2006). On the Benjamini-Hochberg method. {\it The Annals of Statistics}. \textbf{34} 1827-–1849.

\bibitem{2}FIRTH, D. (1993). Bias reduction of maximum likelihood estimates.  {\it Biometrika}.  \textbf{80} 27--38.

\bibitem{2} FONG, L. Y. Y., FIDANZA, V., ZANESI, N., LOCK, L. F., SIRACUSA, L. D., MANCINI, R., SIPRASHVILI, Z.,  OTTEY, M., MARTIN, S. E., DRUCK, T., MCCUE, P. A., CROCE, C. M., and HUEBNER, K. (2000).  Muir–Torre-like syndrome in Fhit-deficient mice.  {\it Proc. Nat. Acad. Sci}. \textbf{97} 4742–-4747.

\bibitem{2} FOSTER, D.P. and STINE, R.A. (2004). Variable selection in data mining: building a predictive model for bankruptcy.  {\it JASA}. \textbf{99} 303--313.

\bibitem{2} FREEDMAN, D. A. (2008). Randomization Does Not Justify Logistic Regression. {\it Statistical Science}.  \textbf{23} 237--249.

\bibitem{2} FRIEDMAN, J. H. (2001).  Greedy function approximation: a gradient boosting machine.  {\it The Annals of Statistics}. \textbf{29} 1189--1232.

\bibitem{2} GUYON, I. and ELISSEE
FF, A. (2003). An introduction to variable and feature selection.  {\it J. of Machine Learning Research}. \textbf{3} 1157--1182.

\bibitem{2} HAND. D. J. (2008). Breast cancer diagnosis from proteomic mass spectrometry data: a comparative evaluation.  {\it Statistical Applications in Genetics and Moecular Biology}. \textbf{7} article 15.

\bibitem{2} HASTIE, T., FRIEDMAN, J. H., and TIBSHIRANI, R. (2011). {\it The Elements of Statistical Learning}.  Springer-Verlag.

\bibitem{2} HEINZE, G. and SCHEMPER, M. (2002). A solution to the problem of separation in logistic regression. {\it Statistics in Medicine}. \textbf{21} 2409--2419.

\bibitem{2} HOLLAND, P. W. (1986). Statistics and causal inference. {\it JASA}. \textbf{81} 945--960.

\bibitem{2} HU, Q., PAN, W., AN, S., MA, P., and WEI, J. (2010). An efficient gene selection technique for cancer recognition based on neighborhood mutual information.  {\it Int. J. Machine Learning \& Cyber}. \textbf{1} 63--74.

\bibitem{2} HUANG, J., BREHENY, P., and MA, S. (forthcoming). A selective review of group selection in high dimensional models. {\it Statistical Science}. (\url{http://www.imstat.org/sts/future_papers.html}).

\bibitem{2} HUANG, J., MA, S., LI, H. Z. and ZHANG, C. H. (2011). The sparse Laplacian shrinkage estimator for high-dimensional regression. {\it The Annals of Statistics}. \textbf{39} 2021-–2046.

\bibitem{2} JEANMOUGIN M, DE REYNIES A, MARISA L, PACCARD C, NUEL G, et al. (2010).  Should we abandon the t-test in the analysis of gene expression microarray data: a comparison of variance modeling strategies.  {\it PLoS ONE}. \textbf{5} e12336.

\bibitem{2} LEE, Y. J., CHANGY, C. C. and CHAO, C. H. (2008).  Incremental forward feature selection with application to microarray gene expression data.  {\it Journal of Biopharmaceutical Statistics}.  \textbf{18} 827--840.

\bibitem{2} LEEK, J. T. and STOREY, J. D. (2011). The joint null criterion for multiple hypothesis tests. {\it Statistical Applications in Genetics and Molecular Biology}.  \textbf{10} article 28.

\bibitem{2} LETTRE, G, PALMER, C. D., YOUNG, T., EJEBE, K. G., ALLAYEE, H, et al. (2011). Genome-wide association study of coronary heart disease and its risk factors in 8,090 African Americans: The NHLBI CARe Project. {\it PLoS Genet}. \textbf{7} e1001300.

\bibitem{2} LIU, J., JI, S., and YE, J. (2009). SLEP: Sparse learning with efficient projections. Arizona State University. (\url{http://www.public.asu.edu/~jye02/Software/SLEP}).

\bibitem{2} MA, P. and WEI, J. M. (2010). An efficient gene selection technique for cancer recognition based on neighborhood mutual information. Int. {\it J. Mach. Learn. \& Cyber}.  \textbf{1} 63-–74.

\bibitem{2} MA, S., SONG, X. and HUANG, J. (2007). Supervised group Lasso with applications to microarray data analysis. {\it BMC Bioinformatics}.   \textbf{8} 1186--1471.

\bibitem{2} MAGIDSON, J. (2010). Correlated component regression: a prediction/classification methodology for possibly many features,” {\it Proceedings of the 2010 Joint Statistical Meeting}. (\url{http://statisticalinnovations.com/technicalsupport/CCR.AMSTAT.pdf}).


\bibitem{2} MONGAN, M.A., DUNN II, R.T. et al. (2010).  A novel statistical algorithm for gene expression analysis helps differentiate pregnane X receptor-dependent and independent mechanisms of toxicity. {\it PLoS One}. \textbf{5} e15595.
 		
\bibitem{2} NAIK, P. A., HAGERTY, M. R., and TSAI, C. L. (2000). A new dimension reduction approach for data-rich marketing environments: sliced inverse regression. {\it J. of Marketing Research}.  \textbf{37} 88--101.

\bibitem{2} PARK, M. Y. and HASTIE, T. (2008). Penalized logistic regression for detecting gene interactions. {\it Biostatistics}. \textbf{9} 30--50.

\bibitem{2} PEREIRA, B. and RAO. C. R. (2009). \textit{Data mining using neural networks: a guide for statisticians}.

\bibitem{2} PHENIX H., MORIN K., BATENCHUK C., PARKER J., ABEDI V., et al. (2011). Quantitative epistasis analysis and pathway inference from genetic interaction data. {\it PLoS Comput Biol}. \textbf{7} e1002048.

\bibitem{2} RAO, K. N., NAGIREDDY, S., and CHAKRABARTI, S. (2011). Complex genetic mechanisms in glaucoma: an overview.   {\it Indian J. Ophthalmol}. \textbf{59} 31--42.

\bibitem{2} RITCHIE, M. D., HAHN, L. W., ROODI, N., BAILEY, R., DUPONT, W. D., PARL, F. F. and MOORE, J. H.  (2001). Multifactor-dimensionality reduction reveals high-order interactions among estrogen-metabolism genes in sporadic breast cancer.  {\it Amer. J. Human Genet}. \textbf{69} 138--147.

\bibitem{2} SHMUELI, G. (2010). To explain or to predict? {\it Statistical Science}. \textbf{25} 289--310.

\bibitem{2} SIERRA, A. and ECHEVERRIA, A. (2003).  Skipping Fisher’s criterion.  {\it Lecture Notes in Computer Science}. 962–-969. Springer-Verlag.

\bibitem{2} SINGH, D., FEBBO, P. G., ROSS, K. et al. (2002). Gene expression correlates of clinical prostate cancer behavior. {\it Cancer Cell}. \textbf{1} 203--209.

\bibitem{2} STEEN, K. V. (2011).  Travelling the world of gene-gene interactions.  {\it Briefings in Bioinformatics}.  \textbf{13} 1--19.

\bibitem{2} STIGLER, S. M. (2010). The changing history of robustness.  {\it The American Statistician}. \textbf{64} 277--281.

\bibitem{2} STOKES, H. H. (2004). On the advantage of using two or more econometric software systems to solve the same problem. {\it J. Econ \& Soc Measurement}. \textbf{29} 307--320.

\bibitem{2} STOREY, J. D.  (2010). False discovery rates. {\it International Encyclopedia of Statistical Science}. 504--508. Lovric, M. (editor). Springer.

\bibitem{2} SU, Y., MURALI, T. M., PAVLOVIC, V. et al. (2003). RankGene: identification of diagnostic genes based on expression data. {\it Bioinformatics Applications}. \textbf{19} 1578-–1579.

\bibitem{2} VINZI, V. E., CHIN, W. W., HENSELER, J., WANG, H. (2010), Eds., Handbook of Partial Least Squares, Springer.
WANG, L., LIU, X., LIANG, H. AND CARROLL, R. J. (2011). Estimation and variable selection for generalized additive partial linear models. {\it The Annals of Statistics}. \textbf{39} 1827–-1851.

\bibitem{2} WANG, X. S. and SIMON, R. (2011). Microarray-based Cancer Prediction Using Single Genes. {\it BMC Bioinformatics}. \textbf{ 12} 391.

\bibitem{2} WOLD, S, SJÖSTRÖM, M., and ERIKSSON, L. (2001). PLS-regression: a basic tool of chemometrics. {\it Chemometrics and Intelligent Laboratory Systems}. \textbf{58} 109-–130.

\bibitem{2} YOUNG, S. S. (2008).  Everything is dangerous. \newline (\url{http://www.niss.org/sites/default/files/Young_Safety_June_2008.pdf}).

\bibitem{2} YOUNG, S. S. and KARR, A. (2011). {\it Significance}. \textbf{8} 116--120.

\bibitem{2} YUAN, M. and LIN, Y. (2007). On the non-negative garrotte estimator. {\it J. R. Statist. Soc. B}. \textbf{69} 143-–161.

\bibitem{2} ZOU, H. and HASTIE, T. (2007). Regularization and variable selection via the elastic net. Department of Statistics, Stanford University.  A preprint.

\bibitem{2} ZUBER, V. and STRIMMER, K. (2011). High-dimensional regression and variable selection using CAR scores.  {\it Statistical Applications in Genetics and Molecular Biology}. \textbf{10} article 34.


\end{thebibliography}
\end{document}